\documentclass[twoside]{article}

\usepackage{cite}

\usepackage[utf8]{inputenc} 
\usepackage[sc]{mathpazo} 
\usepackage[T1]{fontenc} 
\usepackage{microtype} 

\usepackage{multirow}
\usepackage{topcapt}

\usepackage{multicol}

\usepackage{algorithm2e}
\makeatletter
\renewcommand{\@algocf@capt@plain}{above}
\makeatother

\usepackage[hmarginratio=1:1,top=32mm,columnsep=20pt]{geometry} 
\usepackage[hang, small,labelfont=bf,up,textfont=it,up]{caption} 
\usepackage{booktabs} 

\usepackage{lettrine} 

\usepackage{enumitem} 
\setlist[itemize]{noitemsep} 

\usepackage{abstract} 

\usepackage{titlesec} 
\titleformat{\section}[block]{\large\scshape\centering}{\thesection.}{1em}{} 
\titleformat{\subsection}[block]{\large}{\thesubsection.}{1em}{} 

\usepackage{graphicx}
\usepackage{subfig}
\usepackage{fancyhdr} 
\usepackage{titling} 

\usepackage{hyperref} 

\usepackage{amsmath} 

\usepackage{pifont}

\usepackage{xcolor}

\usepackage{mathtools}

\newif\ifpaper
\papertrue

\newcommand{\VEC}[1]{{\boldsymbol{#1}}}
\newcommand{\pdf}{p.d.f.}

\pretitle{\begin{center}\Huge\bfseries} 
\posttitle{\end{center}} 
\title{Unfolding by Folding: a resampling approach to the problem of matrix inversion without actually inverting any matrix} 
\author{%
  \textsc{Pietro Vischia}\\[1ex] 
\normalsize Institut de recherche en Math\'ematique et Physique,\\ 
\normalsize Universit\'e catholique de Louvain\\
\normalsize \href{mailto:pietro.vischia@cern.ch}{pietro.vischia@cern.ch}\\ 
}
\date{\today} 
\renewcommand{\maketitlehookd}{%
\begin{abstract}
  \noindent Matrix inversion problems are often encountered in experimental physics, and in particular in high-energy particle physics, under the name of \textit{unfolding}.
  The true spectrum of a physical quantity is deformed by the presence of a detector, resulting in an observed spectrum.
  If we discretize both the true and observed spectra into histograms, we can model the detector response via a matrix. Inferring a true spectrum starting from an observed spectrum requires therefore inverting the response matrix.
  Many methods exist in literature for this task, all starting from the observed spectrum and using a simulated true spectrum as a guide to obtain a meaningful solution in cases where the response matrix is not easily invertible.

  In this Manuscript, I take a different approach to the unfolding problem.
  Rather than inverting the response matrix and transforming the observed distribution into the most likely parent distribution in generator space, I sample many distributions in generator space,
  fold them through the original response matrix, and pick the generator-level distribution that yields the folded distribution closest to the data distribution.
  Regularization schemes can be introduced to treat the case where non-diagonal response matrices result in high-frequency oscillations of the solution in true space,
  and the introduced bias is studied.

  The algorithm performs as well as traditional unfolding algorithms in cases where the inverse problem is well-defined in terms of the discretization of the true and smeared space,
  and outperforms them in cases where the inverse problem is ill-defined---when the number of truth-space bins is larger than that of smeared-space bins.
  These advantages stem from the fact that the algorithm does not technically invert any matrix and uses only the data distribution as a guide to choose the best solution.

  \end{abstract}
}

\begin{document}

\maketitle

\section{Introduction}

Matrix inversion problems have an important application in high-energy experimental particle physics (HEP):
we accelerate beams of particles at the highest energies achieved by humankind, we make the beams collide
and record the output given by the large particle detectors we place around the beam collision point.
We then typically categorize the events according to the range of values of some observable quantity,
obtaining \textit{histograms} which represent the experimental distribution of that observable.
In order to compute appropriate observables, we convert the raw output of the detector elements (mainly composed by electric signals)
into a set of \textit{reconstructed objects} using several complex algorithms which aim at reconstructing the details of the passage of
a particle through the detector.

We assume that the reconstructed distribution is the expression of an a-priori \textit{true distribution}
that is defined only by the underlying physics of the collision.
In this picture, we can then interpret the detector and event reconstruction as an operator which smears
the true distribution transforming it into the reconstructed distribution.
We model this mathematically by saying that the true distribution $\VEC{x}$ is transformed (\textit{folded} into the
reconstructed distribution $\VEC{y}$ by the response of the detector seen as a \textit{response matrix} $R$, such that
\begin{equation}
  \label{eq:fold}
  \VEC{y}=R\VEC{x}\,.
\end{equation}
We have direct experimental access only to the reconstructed distribution.
Using a Monte Carlo simulation of the physics processes that happen in the collision,
we obtain also a true distribution. We give the simulated events as an input to a
detailed detector simulation, obtaining a simulated reconstructed distribution:
we can then build the response matrix as the two-dimensional histogram of the the simulated true and reconstructed distributions.
The response matrix will crucially depend on the details of the Monte Carlo simulation of the given physics process.

Given a response matrix and an observed spectrum from data taken in real collisions, we can think of transforming the observed spectrum from the reconstructed-level space to the true-level space by inverting Equation~\ref{eq:fold}
to obtain an estimate $\VEC{\hat{x}}$ of the true spectrum corresponding to an observed spectrum $\VEC{y}$:
\begin{equation}
  \label{eq:unfold}
  \VEC{\hat{x}} = R^{-1}\VEC{y}\,.
\end{equation}
Crucially, the process of transforming back (\textit{unfolding}) the reconstructed spectrum to the true one involves the inversion of the response matrix;
a certain level of arbitrariness is involved in the procedure when the matrix is not easily invertible.

Several algorithms have been proposed in literature to solve this problem, all assuming that we need to start from the reconstructed level spectrum and to invert the response matrix in a robust way to obtain a reliable estimate of the true spectrum.
Simple bin-by-bin inversion is always discouraged, because it does not take into account any migration effect.
The analytic inversion of the response matrix is sometimes not feasible, and can be problematic even when it is feasible.
The inverted matrix will crucially depend on the reciprocal of the determinant of the original matrix, and when the matrix is not diagonal the determinant may be close to zero;
its reciprocal may then be very large, and the reconstructed spectrum may be mapped into an unfolded distribution affected by high-frequency fluctuations with large amplitude, as shown by Schmitt~\cite{Schmitt:2016orm}.

To solve the probem of high-frequency fluctuations, an artificial constraint is introduced as a way of \textit{regularizing} the problem;
the variance of the unfolded solution is severely reduced, at the price of introducing a bias towards the true spectrum assumed as the target for the regularization term.
It is possible to study the bias, but nevertheless regularization should be used only when strictly necessary,
preferring unregularized methods whenevery possible.
In particular, when the response matrix is almost diagonal, introducing a regularization term does not improve the measurement.

Classical unfolding methods include a simplification of a maximum likelihood estimate by performing a $\chi^2$ minimization.
The popular \textsc{TUnfold} software by Schmitt~\cite{Schmitt:2012kp} solves the matrix inversion by minimizing the sum
of lagrangian terms corresponding to the raw inversion of the response matrix, to the correction of the $\chi^2$ Gaussian assumption
via a constraint to the normalization of the data distribution, and to regularization.
The regularization lagrangian consists in an explicit Tikhonov term~\cite{tikhonov} characterized by
a \textit{bias vector} which gives the target for the smoothing of fluctuations, a metric
which specifies if the smoothing should be applied to the distribution or to its first or second derivatives.
The second derivative (curvature) is often a robust choice.
The strength of the regularization is governed by a parameter called \textit{regularization strength}.
This method is commonly known in statistics literature as ridge regression~\cite{doi:10.1080/00401706.1970.10488634}.
The value of the regularization strength is chosen by scanning some quantity sometimes related to a $\chi^{2}$.
Other algorithms are based on iterating to convergence; regularization is introduced in the form of early-stopping,
as for example is the case for the D'Agostini method that is based on the Bayes theorem and originally~\cite{DAgostini:1994fjx} does not include any iteration procedure; iteration is deemed necessary to reduce effects from the dependence of the posterior on the suggested flat prior.
An improved version is called Iterative D'Agostini method~\cite{DAgostini:2010hil} and consists in careful modelling of some effects
which were approximated with Gaussian distributions in the original version of the algorithm,
and suggests that a careful modelling of the prior fed to the first iteration of the algorithm (flat in the original version)
might help in avoiding or (in practice) at least reducing the need of iterations.
The resulting algorithm is essentially frequentist in that it iterates to convergence to the maximum likelihood solution.
Choudalakis~\cite{2012arXiv1201.4612C} solves the unfolding problem in a fully Bayesian way by sampling the posterior distribution
for the inverse problem with a grid search or with a full-fledged variant of the Metropolis-Hastings algorithm. Regularization is implemented as choice of the prior.
The HEP community has never adopted this algorithm to my knowledge, probably out of concerns
for the severe dependence of the result on the choice of prior.
Glazov~\cite{Glazov:2017vni} has attempted to unfold spectra using machine learning, framing the problem as a classification one;
each reconstructed data event is assigned to the truth bin that most probably originated it, with an iterative procedure.
The true space is discretized before the process starts, thus inducing a regularization of the problem even before unfolding is performed, and coverage tests are not shown.

A variant of the Tikhonov regularization is based on decomposing a square response matrix in its eigenvalues, and choosing
the regularization strength based on the significance of the eigenvalues~\cite{HOCKER1996469}
and is implemented in the \textsc{RooUnfold}~\cite{Adye:2011gm} software.
At the moment of this writing, \textsc{RooUnfold} does not support yet a custom bias vector, so it should be used with great care, if at all.

The choice of the best regularization strength or of the best stopping point for an iterative regularization procedure is quite delicate;
Schmitt~\cite{Schmitt:2016orm} points out that increasing too much the number of iterations can even worsen the result.
Not choosing correctly the number of iterations can induce unwanted and non-studied regularization biases,
and failing to report the number of iterations and the method used to determine it might make the result untrustable. 

The easiness of the matrix inversion depends also on the number of degrees of freedom for the problem,
which is related to the discretization of the true and reconstructed spectra.
Denoting with $N_{bins}(true)$ and $N_{bins}(reco)$ the number of categories (bins) of the
true and reconstructed spectra, respectively, the problem will be ill-defined if $N_{bins}(reco)<N_{bins}(true)$, with equality being the minimal case.

The dangers and the degree of arbitrariness involved in any unfolding procedure based on matrix inversion motivate the usefulness of a change of perspective.

In this Manuscript I take a different approach to the unfolding problem.
Rather than inverting the response matrix and transforming the data distribution into the most likely parent distribution in generator space,
I sample many distributions in generator space, fold them through the original response matrix, and pick the generator-level distribution which yields the folded distribution closest to the data distribution.
Since the choice of best true spectrum is based on proximity to the data, the method does not introduce any bias to the true simulated distribution.
The advantage of this procedure is that the matrix inversion problem is solved without any matrix inversion.
Among the several advantages, this method is therefore insensitive to the amount of bins in the reconstructed and true spectra,
retaining performance in situations where the traditional algorithms have the most issues, namely when $N_{bins}(reco)<N_{bins}(true)$.
Cases where the response matrix is severely non-diagonal still result in ill-posedness that requires regularization methods.

I have \textit{arXived} this Manuscript to make it coincide with a talk I gave~\cite{vischiaIcnfp2020} on this algorithms at the ICNFP 2020 conference in the sunny Crete.
This may have resulted in having too hastily drafted some sections and lacking some further tests of the proposed method (such as coverage tests): for that I apologize to the reader.
A second version of this draft will, in due time, contain these checks. I hope the new version will also be enriched by your feedback. Science, after all, is a work in progress.

A template software will be available in due time at the referenced URL~\cite{vischiaAccompanyingSoftware}, containing the full implementation of the the algorithm and the various simulation studies.

\section{Simulation scenarios}
\label{sec:simulation}
I study the performance of the proposed algorithms in three simulated datasets corresponding to four different discreteness scenarios,
depending on the difference between the number of bins of the true spectrum ($N_{bins}(gen)$) and that of the observed spectrum ($N_{bins}(reco)$).

\subsection{Scenarios}
\begin{figure}[!h]
  \centering
  \includegraphics[width=0.49\linewidth]{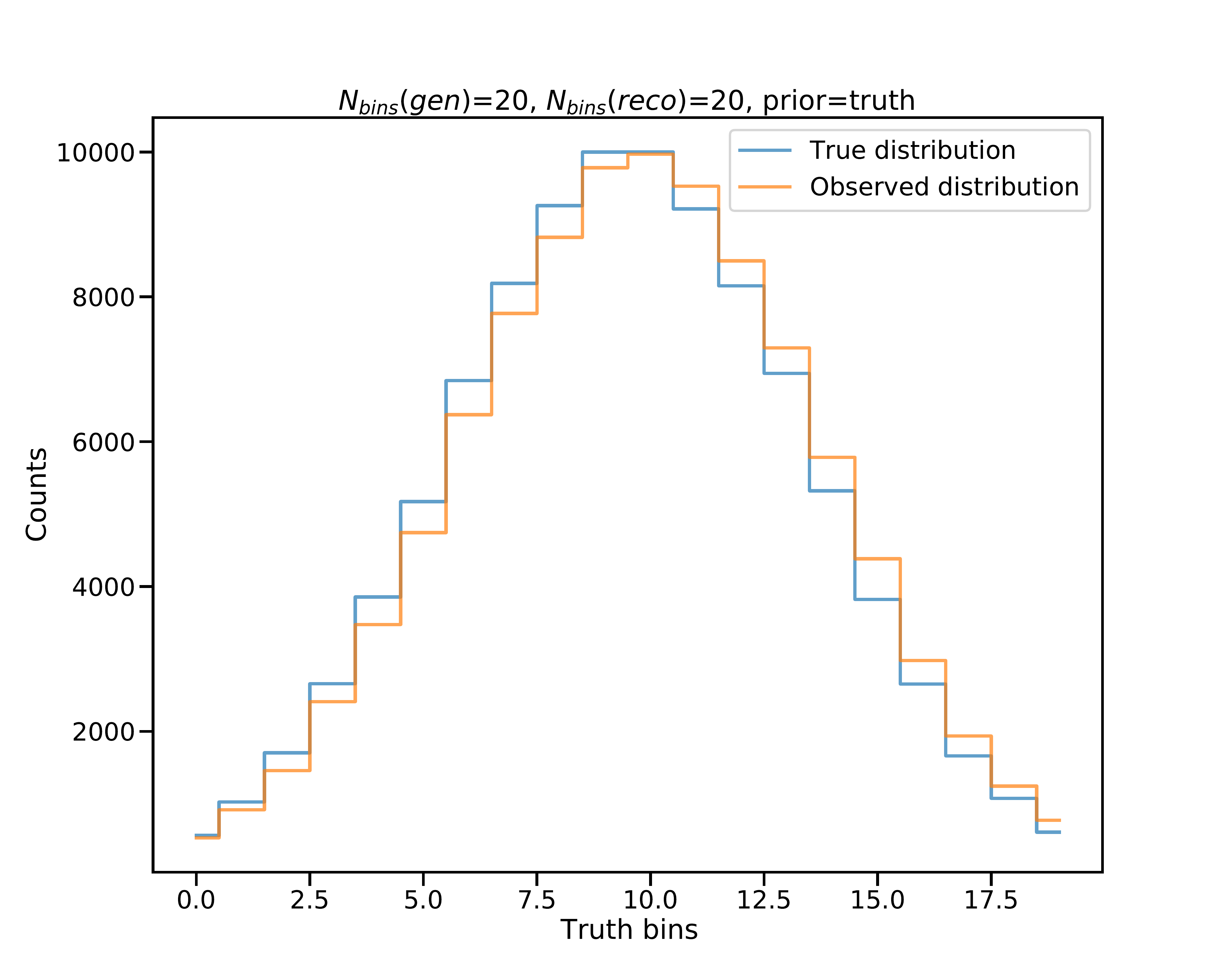}
  \includegraphics[width=0.49\linewidth]{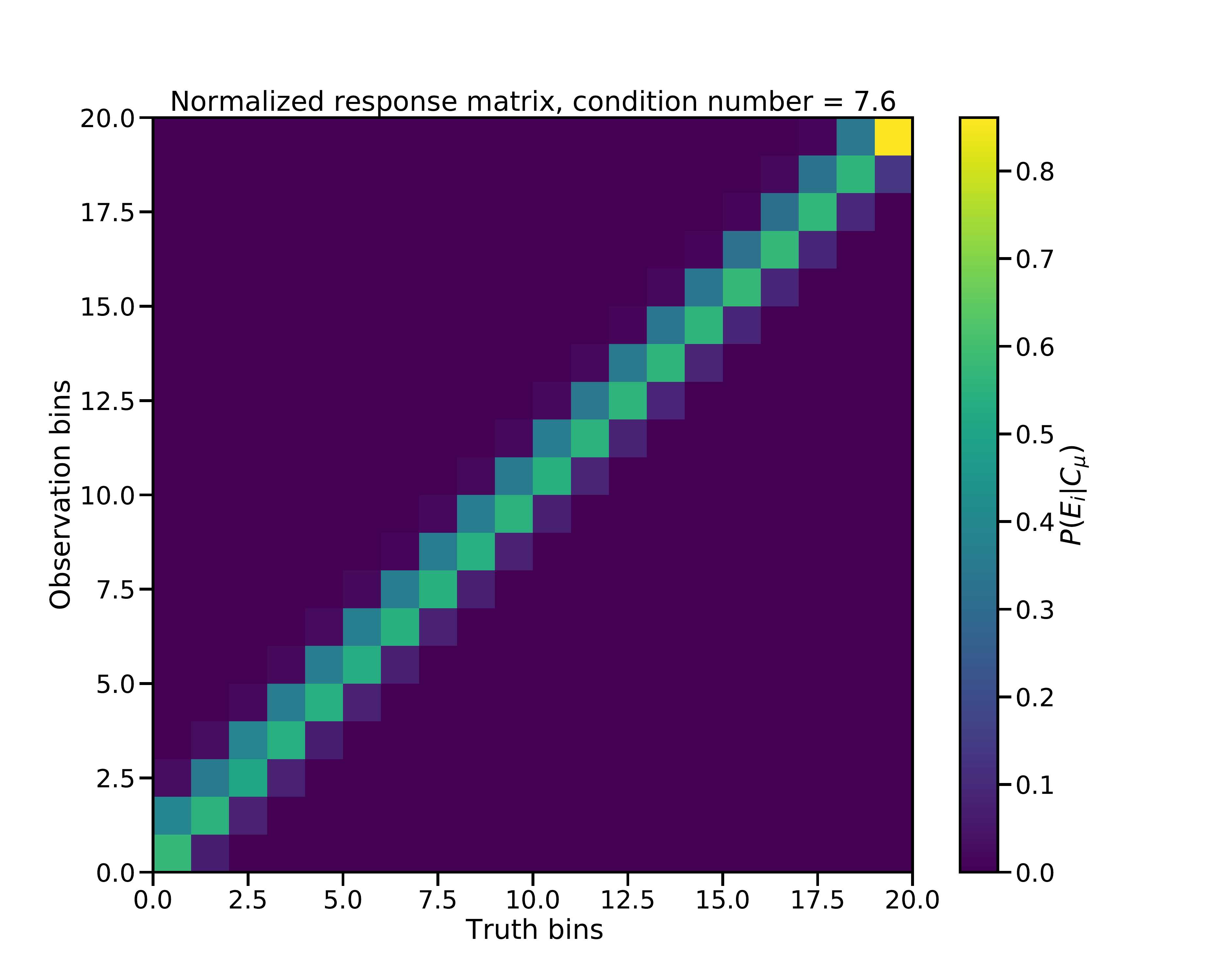}
  \caption{The truth-level and the observation-level distributions (left) for $10^5$ random samples, for $N_{bins}(gen)=20$ and $N_{bins}(reco)=20$. The distributions are used to define the response matrix for the measurement (right), which is normalized across each truth-level bin.}
  \label{fig:resp_reco10gen10}
\end{figure}

The first scenario corresponds to $N_{bins}(gen)=N_{bins}(reco)$ and is characteristic of algorithms like SVD unfolding (the response matrix is square and can be decomposed in singular values);
Without any loss of generality, I discretize the distributions in 20 bins of constant width, $N_{bins}(gen)=N_{bins}(reco)=20$ (Figure~\ref{fig:resp_reco10gen10}, left).

\begin{figure}[!h]
  \centering
  \includegraphics[width=0.49\linewidth]{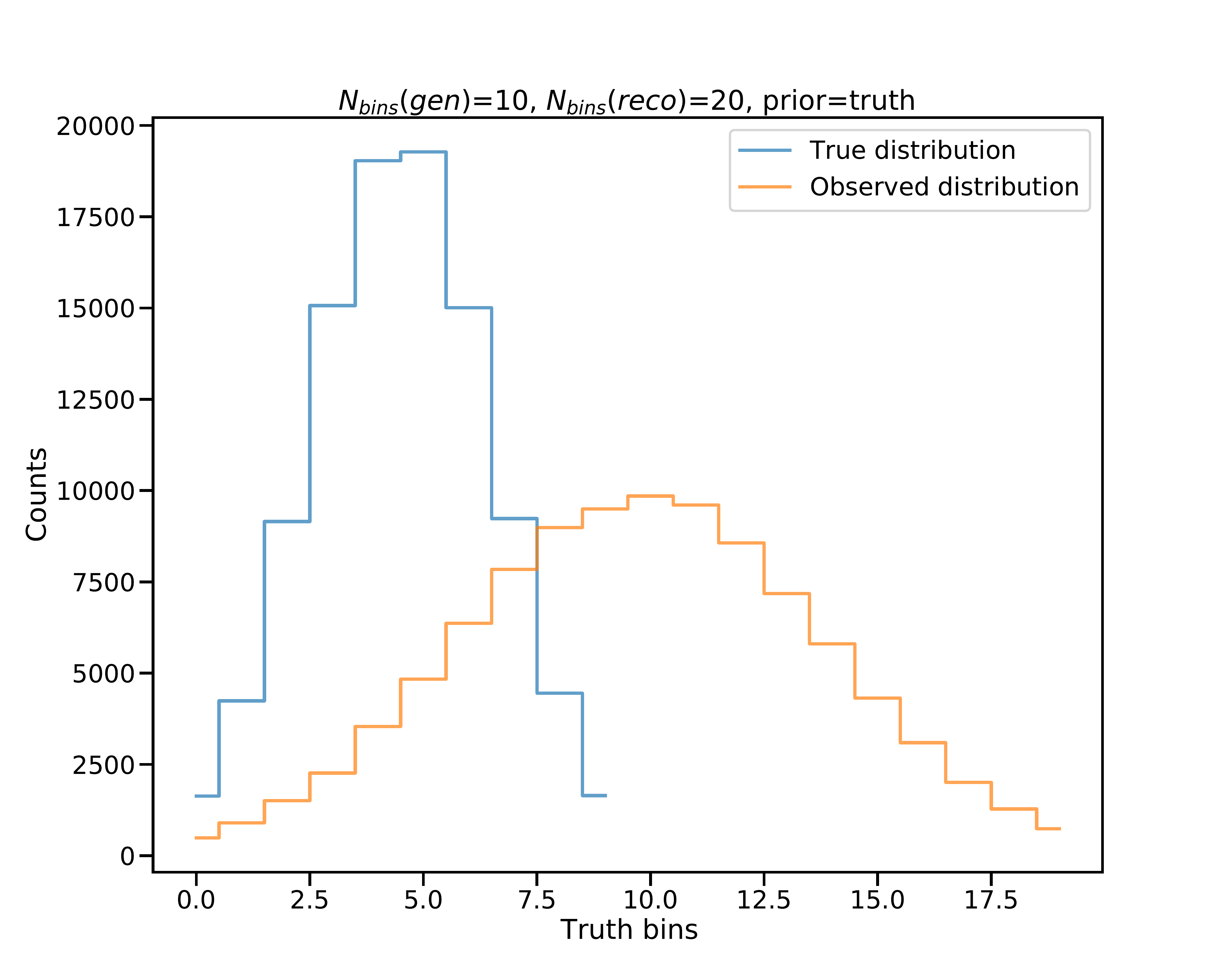}
  \includegraphics[width=0.49\linewidth]{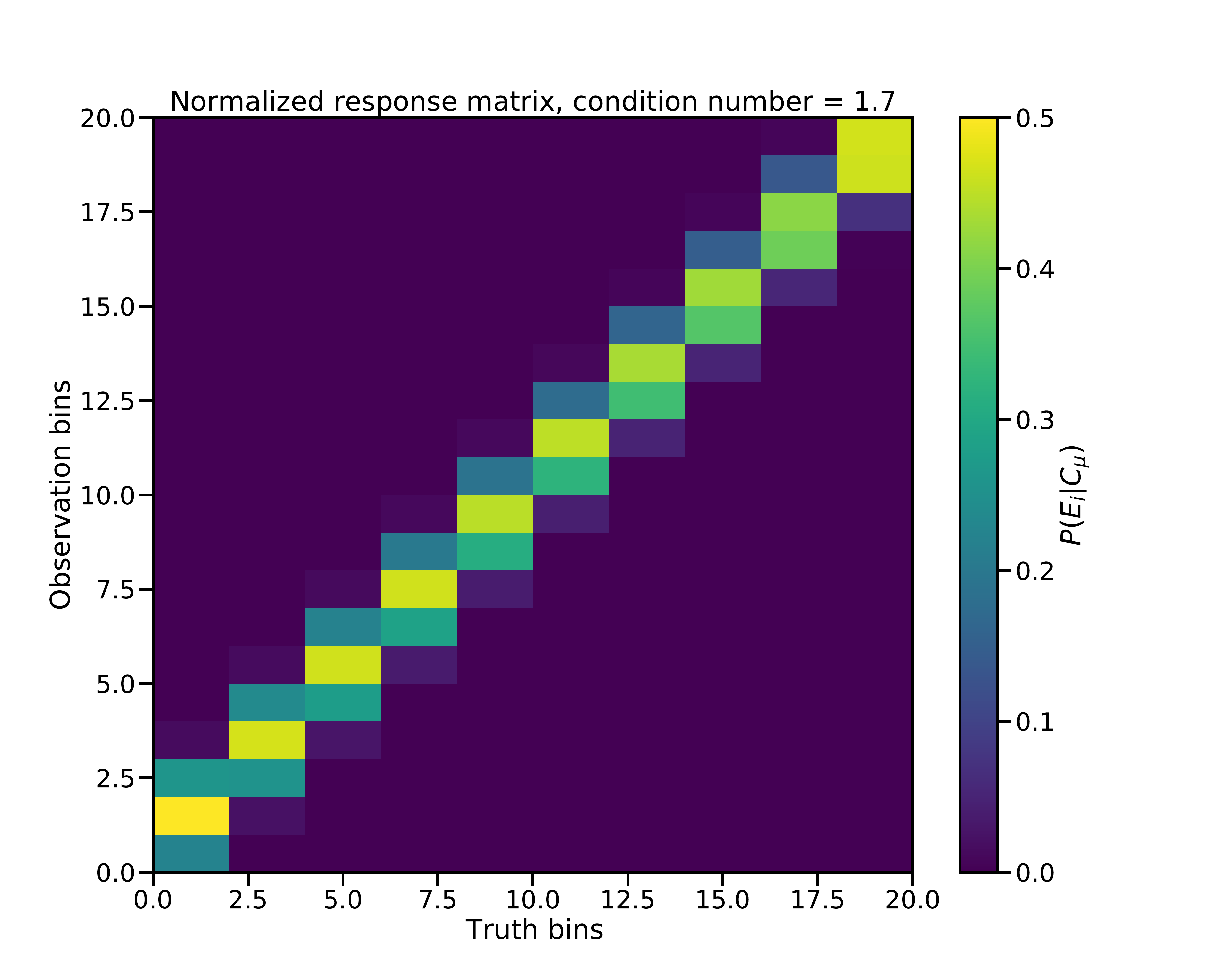}
  \caption{The truth-level and the observation-level distributions (top) for $10^5$ random samples, for $N_{bins}(gen)=10$ and $N_{bins}(reco)=20$. The distributions are used to define the response matrix for the measurement (bottom), which is normalized across each truth-level bin.}
  \label{fig:resp_reco10gen5}
\end{figure}

The second scenario is typical of unfolding algorithms based on $\chi^2$ minimization like \textsc{TUnfold};
these algorithms can in principle work also for $N_{bins}(gen)=N_{bins}(reco)$, but achieve a better performance when $N_{bins}(gen)<N_{bins}(reco)$.
I discretize the distributions in bins of constant width, $N_{bins}(gen)=10$, $N_{bins}(reco)=20$ (Figure~\ref{fig:resp_reco10gen5}, left).

The third scenario corresponds to the case where the inverse problem is ill-defined, $N_{bins}(gen)>N_{bins}(reco)$.
I discretize the distributions into bins of constant width, $N_{bins}(gen)=20$, $N_{bins}(reco)=10$ (Figure~\ref{fig:resp_reco10gen5}, left).

For all scenarios I generate a true spectrum $\VEC{x}$ by sampling $10^5$ times from a gaussian \pdf, $\VEC{x}\sim Gaus(10,4)$,
and a folded observation distribution $\VEC{y}$ by smearing the truth-level distribution with a random noise $\VEC{\epsilon}\sim Gaus(0.3,0.5)$
and the statistical uncertainty associated to Poisson counts.
These distributions are used to build the response matrix, and take the role of Monte Carlo simulation of the process being investigated.

I also generate another pair of distributions: an alternative truth distribution, which takes the role of the real physics process,
and its folded realization, which is taken as the data distribution.
The alternative truth is obtained by adding a random shift $\VEC{\epsilon}\sim Gaus(1.0,0.3)$ to the original truth distribution;
the alternative observed distribution is obtained by smearing the alternative truth with the same random noise that contributes to creating the detector response matrix.

\begin{figure}[!h]
  \centering
  \includegraphics[width=0.49\linewidth]{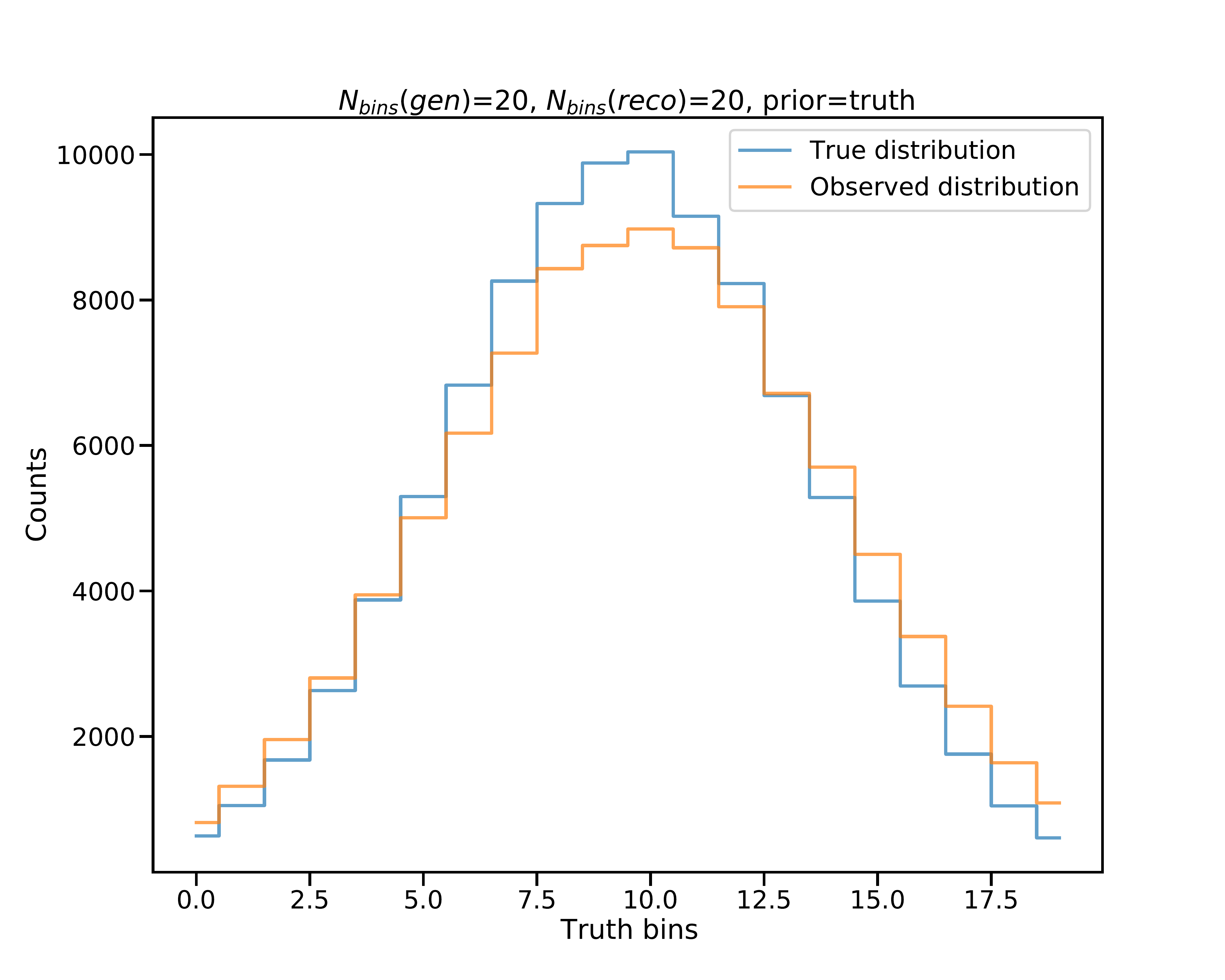}
  \includegraphics[width=0.49\linewidth]{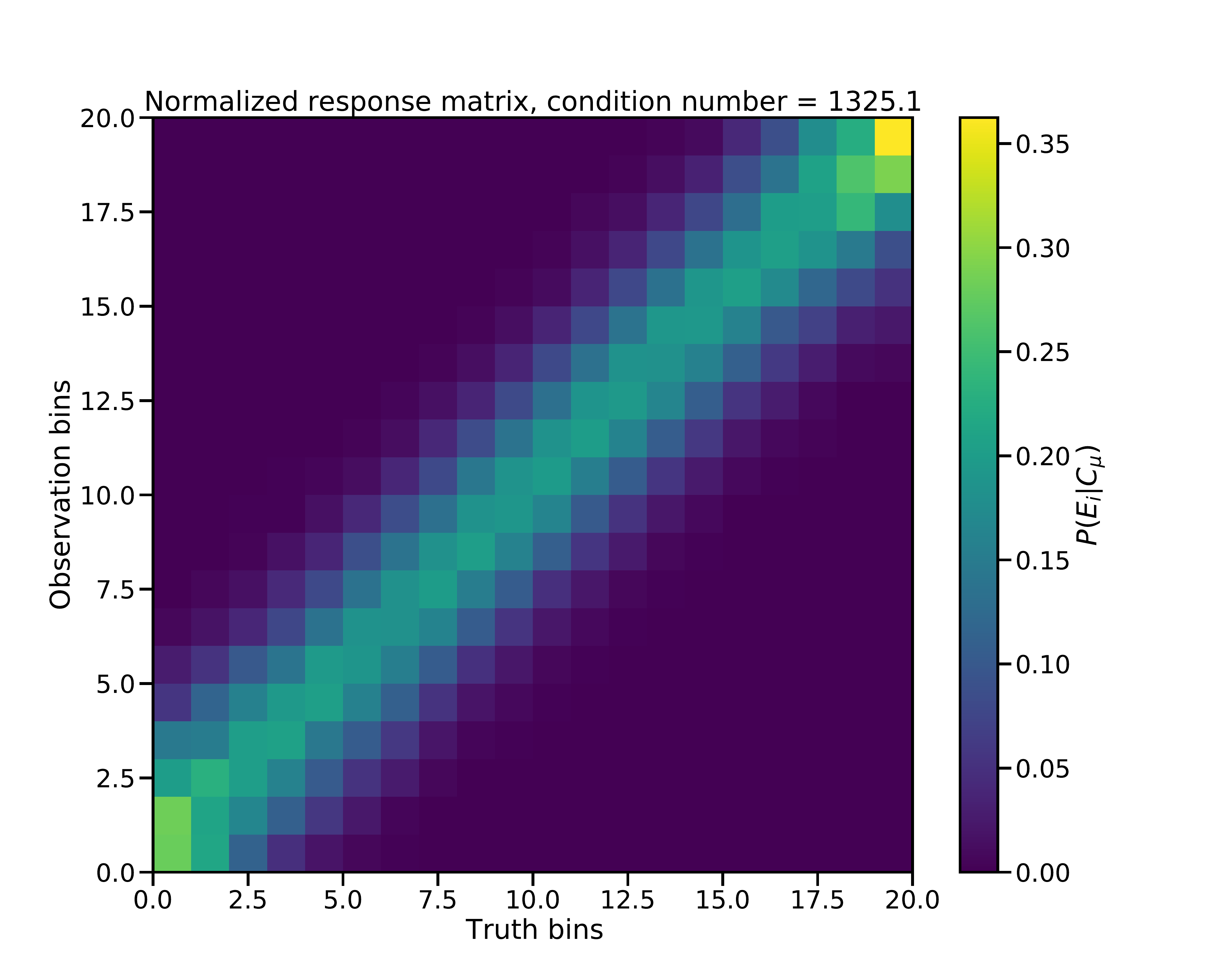}
  \caption{The truth-level and the observation-level distributions (left) for $10^5$ random samples, for $N_{bins}(gen)=20$ and $N_{bins}(reco)=20$, for a remarkably non-diagonal respons. The distributions are used to define the response matrix for the measurement (right), which is normalized across each truth-level bin.}
  \label{fig:resp_NDEXTREMEreco10gen10}
\end{figure}

The fourth scenario corresponds to a case where the response matrix is severely non-diagonal.
For this scenario, the true spectrum is still sampled from a gaussian \pdf $\VEC{x}\sim Gaus(10,4)$,
but the smeared spectrum is obtained by applying a stronger smearing, $\VEC{\epsilon}\sim Gaus(0.3, 2.0)$.
To decouple the effect of ill-definedness coming from the number of true vs smeared bins,
the fourth scenario examines the case where $N_{bins}(gen)=N_{bins}(reco)=20$ (Figure~\ref{fig:resp_NDEXTREMEreco10gen10}).

The situation described by this spectrums of samples is typical in HEP.
We have a simulation for the signal in truth space, and its reconstructed-space version obtained by passing the truth level distribution through a simulation of the detector:
these correspond to the original distributions described above, and are used also to define the response matrix for the unfolding.
The true spectrum from nature might be different though, and result in a data distribution which is different from the one that results from the available theory.
The alternative spectra take therefore the role of the underlying truth and of the data distribution that we would observe in our real detector.
The idea is then to use the original truth distribution as a starting point, the original response matrix as the response matrix, and the alternative data distribution as the data.
A first cross-check is that the algorithm, when run using the original data distribution as target,
identifies back the original truth distribution (which had been used to generate these data), using the response matrix.
The real objective however is that that the algorithm unfolds the alternative data distribution---which is obtained with the alternative truth---using the original truth and the original response matrix.
This would be ideal, in that it would demonstrate that this algorithm does not yield answers biased by the available simulations.

\begin{figure}[!h]
  \centering
  \includegraphics[width=0.49\linewidth]{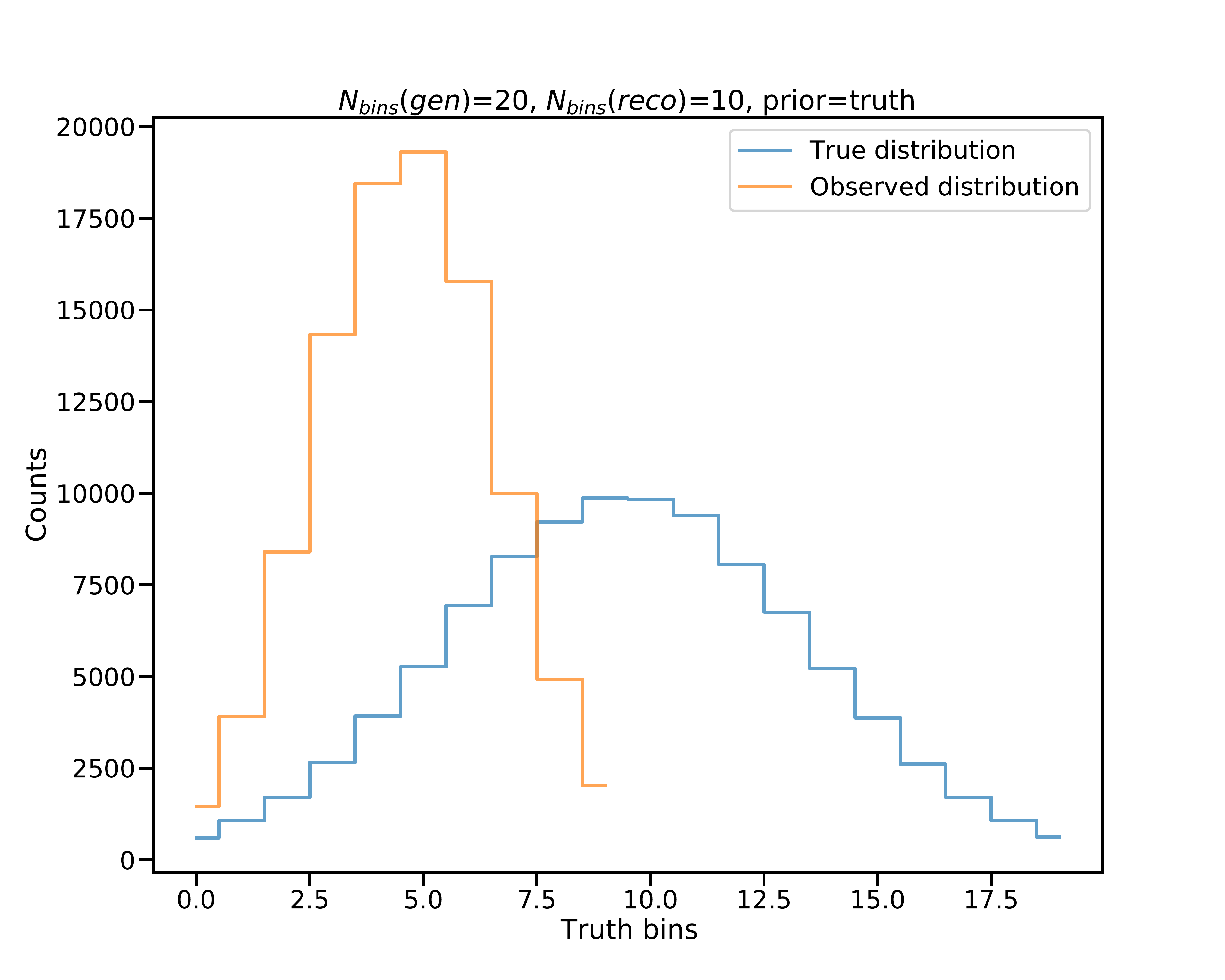}
  \includegraphics[width=0.49\linewidth]{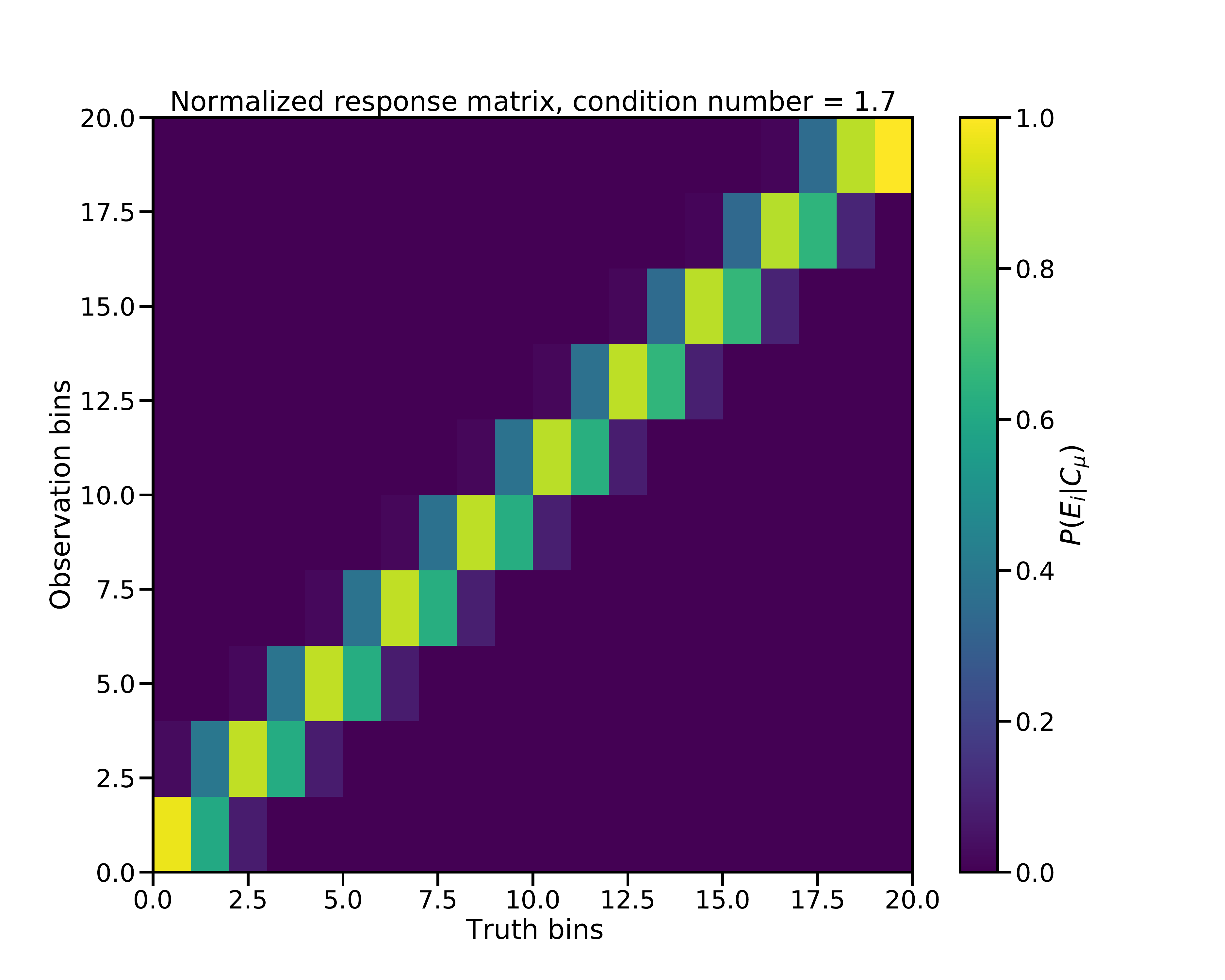}
  \caption{The truth-level and the observation-level distributions (left) for $10^5$ random samples, for $N_{bins}(gen)=20$ and $N_{bins}(reco)=10$. The distributions are used to define the response matrix for the measurement (right), which is normalized across each truth-level bin.}
  \label{fig:resp_reco5gen10}
\end{figure}

The response matrix $R$ is defined by the relationship $\VEC{y} = R\VEC{x}$;
I normalize it such that the probabilities for an individual truth-level bin to migrate to any observation-level bin sum up to 1.
The situation for the three cases is shown respectively in Figures~\ref{fig:resp_reco10gen10} (right),~\ref{fig:resp_reco10gen5} (right), and ~\ref{fig:resp_reco5gen10} (right)
for the three discreteness scenario. The left panels show the generated distributions involved in building the response matrixes.
The diagonality of the response matrix $R$ is assessed by computing their condition number, defined as
\begin{equation}
  cond(R) := \frac{\sigma_{\text{max}}}{max(0, \sigma_{\text{min}})}\,,
\end{equation}
where $\sigma_{\text{max}}$ and $\sigma_{\text{min}}$ are respectively the largest and the smallest singular values in a decomposition of the matrix $R$.
While in principle the condition numbers should be positive definite, the finite precision of floating point calculations can yield negative minimum singular values,
rendering the $max()$ operation at the denominator necessary.
Diagonal matrices are characterized by low condition numbers, $cond(R) = \mathcal{O}(10)$, while severely non-diagonal matrices are characterized by large condition numbers, $cond(R) = \mathcal{O}(10^5)$.
The condition number is displayed in each figure displaying a response matrix.

Each bin of the response matrix is assigned a statistical uncertainty assuming Poisson counts.
Reconstruction efficiency is accounted for (although in the current tests it is set to 100\%).
I obtain two alternative matrices systematic uncertainty by applying a flat 10\% systematic uncertainty to the truth-level distribution and using the shifted distribution to rebuild the response matrix.

\section{Unfolding by folding}
\label{sec:unfoldbyfolding}

\begin{algorithm}[h]
  \SetKwInOut{Input}{input}\SetKwInOut{Output}{output}
  \Input{A truth distribution $\VEC{x}_s$;\\ the data distribution $\VEC{y}$;\\ the response matrix $R$;\\ the number of iterations $N_{samples}$;\\ a statistical test $test$ and an optional condition $oc$ for similarity between histograms;\\ an optional early-stopping condition $sc$.}
  \Output{The random sample $\VEC{\hat{x}_{best}}$ whose folded version matches the better with the data distribution.}
  $\VEC{\hat{x}}_{best}\leftarrow \VEC{x}_s$\;
  $TS_{best}\leftarrow 10^5$\;
  \For{$i \leftarrow 0$ \KwTo $N_{samples}$}{
    $\VEC{x}_i \sim Pois(\VEC{x}_{best})$\;
    $\VEC{y}_i \leftarrow R\VEC{x}_i$\;
    $TS\leftarrow test(\VEC{y}_i, \VEC{y})$\;
    \If{$TS < TS_{best}$ and $oc$}{
      $\VEC{\hat{x}}_{best}\leftarrow\VEC{x}_i$\;
      $TS_{best}\leftarrow TS$\;
    }
    \If{$sc$}{
      \Return
    }
  }
  \caption{The algorithm to resample in the truth-level space until the folded distribution of the sample coincides with the data distribution. I have to put an early stopping and remove the maximum number of iterations.}
  \label{algo:thealgo}
\end{algorithm}

The algorithm resamples from a given distribution in the truth-level space.
The starting distribution can be the truth level distribution, or a flat distribution, or any other distribution.
The resampling is done assuming a Poisson distribution in each bin.
The resulting candidate unfolded truth distribution is multiplied by the response matrix to obtain a candidate reco-level distribution.
The candidate reco-level distribution is compared to the data distribution using Pearson's $\chi^2$, with a stopping criterion $sc$
requiring that reaching a test statistic value of $TS_{Pearson}:=\chi^2/NDOF\geq 0.9$ stops the iteration.
A further early-stopping criterion $oc$ requires that $abs(TS_{Pearson}-1)<0.01$ but it is seldom verified.
Bins with empty reference yield (yield of the observed distribution bin) are merged together with the next non-empty bin.
I discuss alternative statistical tests of compatibility between distribution in Section~\ref{sec:statTest}.
The candidate unfolded truth distribution is then used as a basis for another resampling step.
The procedure is repeated until the folded candidate matches almost perfectly with the data distribution;
An earlier version of this algorithm resampled always from the input truth; the convergence is significantly slower,
and the algorithm is suited only to cases where the smeared truth is already close to the data.
The logic diagram for the algorithm is shown in Algorithm~\ref{algo:thealgo}.

Although there is an unavoidable dependence on the truth distribution encoded in the response matrix,
the algorithm uses only the data to deform the starting candidate until it has a suitable shape. For this reason,
it does not intrinsically bias the result to the truth distribution, as any regularization-based unfolding algorithm would instead do.
A possible issue---highlighted by the results in Section~\ref{sec:simulation}---is that repeating the procedure for the response matrix with varied upwards and downwards fluctuations
provides some constraining power (the target distribution for the resampling with varied response matrix is still the data distribution), but is not expected to seriously outperform other methods.

It turns out that the idea of unfolding by sampling in true space has been proposed by Landweber~\cite{10.2307/2372313}, who suggested using a simple gradient descent to find the best solution.
The method presented in this paper can therefore be considered a stochastic-optimization version of the Landweber method.

\begin{algorithm}[h]
  \SetKwInOut{Input}{input}\SetKwInOut{Output}{output}
  \Input{The truth level distribution $\VEC{x}$ or a generic \pdf;\\ the reco-level data distribution $\VEC{y}$;\\ the response matrix with full dependence on the nuisance parameters $R(\alpha)$;\\ the number of iterations $N_{samples}$}
  \Output{The random sample $\VEC{\hat{x}_{best}}$ whose folded version matches the better with the data distribution}
  $\VEC{\hat{x}}_{best}\leftarrow \VEC{x}$ (or a generic \pdf)\;
  $\chi^2_{best}\leftarrow 10^5$\;
  \For{$i \leftarrow 0$ \KwTo $N_{samples}$}{
    $\VEC{x}_i \sim Pois(\VEC{x}_{best})$\;
    $\VEC{y}_i \sim R(\alpha)\VEC{x}_i$\;
    $c\leftarrow\chi^2_{NDOF}(\VEC{y}_i, \VEC{y})$\;
    \If{$c < \chi^2_{best}$ and $c\geq1$}{
      $\VEC{\hat{x}}_{best}\leftarrow\VEC{x}_i$\;
      $\chi^2_{best}\leftarrow c$\;
    }
  }
  \caption{The algorithm to resample in the truth-level space until the folded distribution of the sample coincides with the data distribution. I have to put an early stopping and remove the maximum number of iterations.}
  \label{algo:algoabc}
\end{algorithm}

Algorithm~\ref{algo:algoabc} increases an analogy to Approximate Bayesian Computation (ABC) by embedding the estimate of the systematic uncertainties in the resampling step.
The folding step is performed by sampling from the full \pdf for the response matrix, including the systematic uncertainties,
and a full ``posterior'' for the unfolded distribution (one distribution per each bin in truth space) is obtained;
the central estimate and the uncertainty bands are obtained by taking the appropriate quanties of this posterior distribution.
This procedure is similar to the ABC$\mu$ proposed by Ratmann and colleagues~\cite{Ratmann10576},
except that here there is no need to assume a simplified likelihood (the model for the response matrix and its uncertainties is assumed as known).
It would be possible to use ABC$\mu$ to introduce an additional uncertainty term to cover for unknown systematics;
while this is certainly interesting to apply the Unfolding by Folding method in other disciplines,
in the context of HEP this is not considered a good practice because, as Cousins~\cite{Cousins:2013hry} points out,
we usually assume that the Standard Model reasonably reflects the underlying laws of nature.
The possibility should not be easily dismissed, however, because as Box~\cite{10.2307/2286841} very rightly notes, \textit{``all models are wrong''}.
Simulation studies for this alternative algorithm are ongoing.

It should be highlighed that building the class of solutions that can be used to estimate the uncertainty is a heavily parallelizable process.

\subsection{Test of equality between the folded histogram and the observations}
\label{sec:statTest}

There is no universally-optimal test for consistency of two histograms, as highlighted by Porter~\cite{Porter:2008mc}. A similar conclusion can be drawn from Thaas~\cite{thaasbook}

The basic version of the algorithm I propose minimizes the usual Pearson $TS_{Pearson}$;
the modification by Porter $TS_{\chi^2}^{Porter}= \sum_{i=1}^{n} \frac{(p_i-q_i)^2}{p_i+q_i}$ perform equally well,
provided that---as for $TS_{Pearson}$---the bins with empty reference yield are merged together with non-empty bins.
The full \textit{minbin} proposal by Porter of merging bins until a configurable minimum yield per bin is reached does not significantly improve the performance of the algorithm;
this is because the algorithm iterates until the residuals between the two histograms are essentially null.
The Bhattacharyya distance~\cite{bdm} is found to work similarly well.
It must be noted that the reference distribution in this case is the data distribution, which is fixed;
the folded distribution for the $i$-th random sample in truth space is the probe whose distance from the data is minimized.

A similar performance (both in terms of speed of convergence and of quality of the result) can be obtained by using a Wasserstein $p$-distance with $p=1$ (i.e. the Earth Mover distance)~\cite{dezadeza}.
I use the distribution-free version described by Ramdas and colleagues~\cite{2015arXiv150902237R}.

Entropy-based test statistics like the Kullback-Leibler (KL) divergence~\cite{sivia,bda3} results in a computationally more intensive algorithm,
which captures very well the mean of the distribution.
However, the minimization of the KL divergence results in histograms which have a compatible mean from the point of view of information theory;
we are instead interested in point-by-point equality of the histograms.
Furthermore, some forms of divergence distance are not suitable to discrete cases,
because they involve a difference between bin-by-bin yields at the denominator, which would be zero with significantly non-zero probability.

A class of algorithms proposed by Diakonikolas and colleagues~\cite{2017arXiv170301913D} are interesting but probably not too useful:
in their formulation, each statistical test requires heavy resampling from one of the two distributions,
but Algorithm~\ref{algo:thealgo} performs several hundred thousands (millions, in the more difficult cases) tests and it is therefore unfeasible to resample within each test.

\subsection{Application of the algorithms}
Algorithm~\ref{algo:thealgo} is then used to generate a prediction corresponding to the best folded distribution when compared to the observation-level distribution.
The algorithm is also used to generate alternative predictions by using the up and down systematic variations of the response matrix.
I also perform a D'Agostini unfolding iterating to convergence and use the results as a reference to assess the quality of Algorithm 1.
For the first two scenarios (well-defined inverse problem) I run $10^5$ iterations,
whereas the convergence is slower for the third scenario (less observed bins than truth ones) and the algorithm requires about $10^6$ iterations to converge.
The full algorithm, including iterative convergence for the systematic uncertainties, requires anyways a handful minutes to run.

\begin{figure}[!h]
  \centering
  \includegraphics[width=0.49\linewidth]{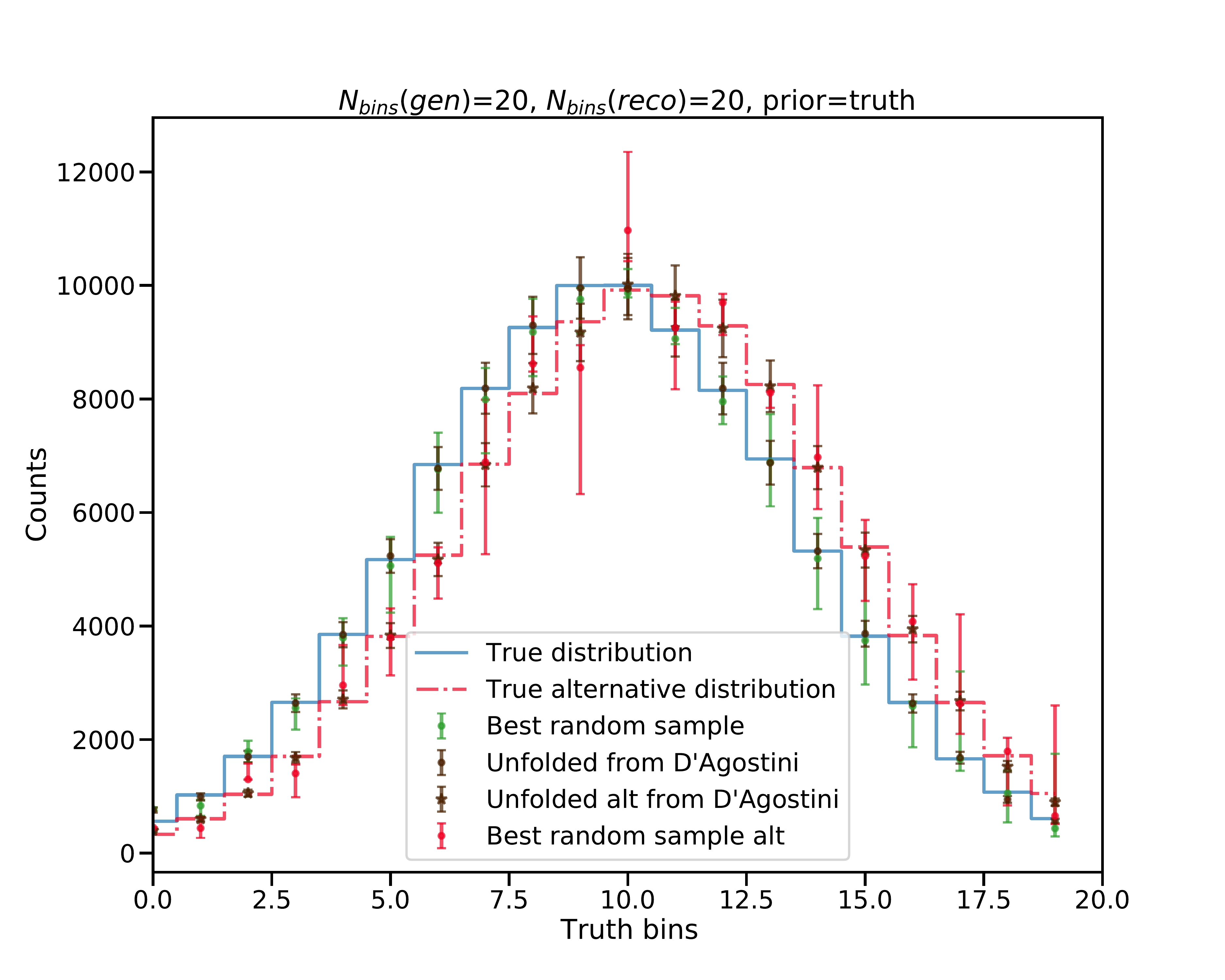}
  \includegraphics[width=0.49\linewidth]{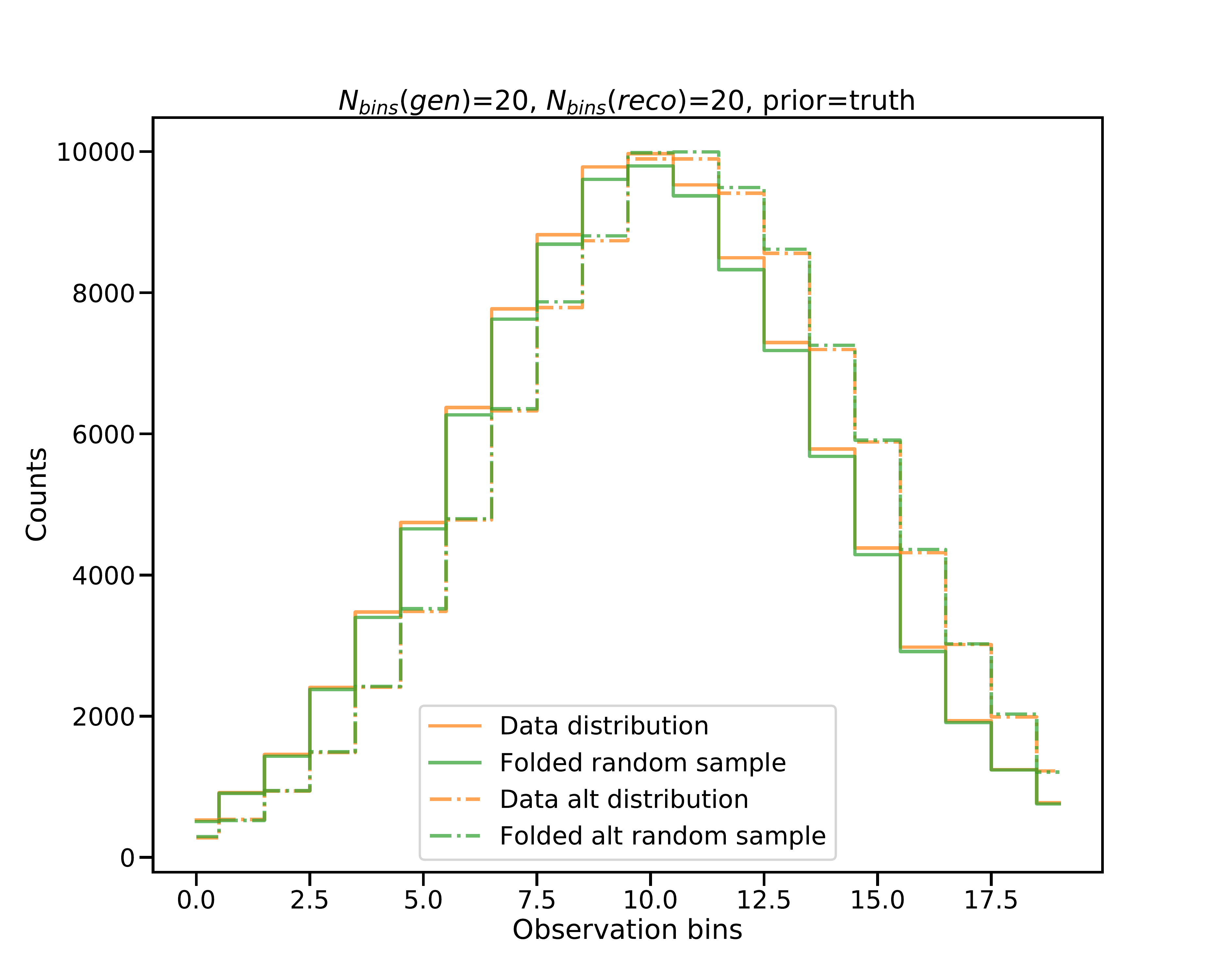}
  \caption{The truth-level (left) and observation-level (right) distributions together with the results of the resampling algorithms, for $N_{bins}(gen)=20$ and $N_{bins}(reco)=20$. In the truth-level space, the true distribution is shown together with the random sample that yields the best comparison with the observation-level distribution. The unfolded distribution using the iterative D'Agostini method is also shown. In the observation-level space, the smeared data distribution and the folded random sample distribution that yields the best comparison with the observation-level distribution.
  }
  \label{fig:fold_reco10gen10}
\end{figure}

\begin{figure}[!h]
  \centering
  \includegraphics[width=0.49\linewidth]{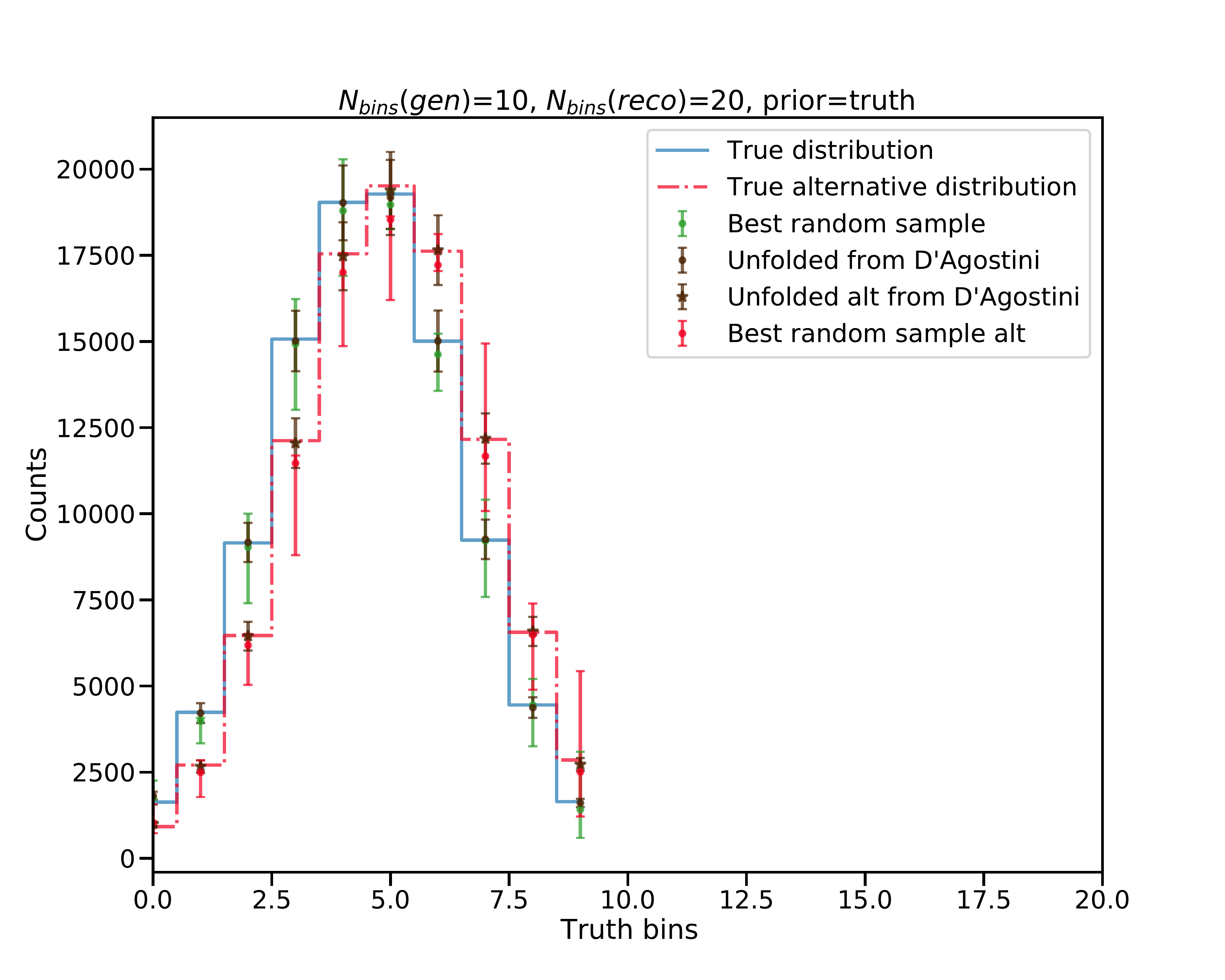}
  \includegraphics[width=0.49\linewidth]{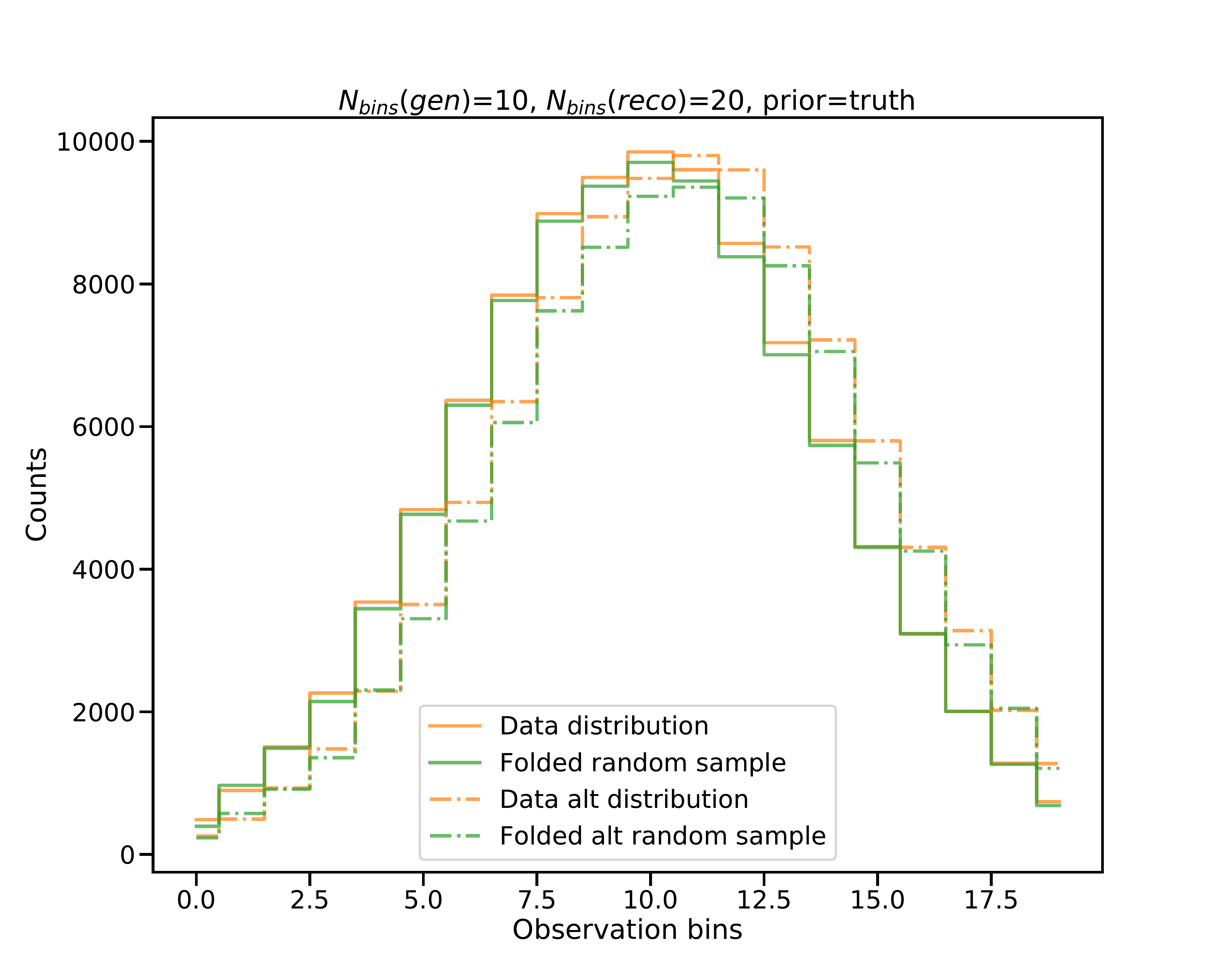}
  \caption{The truth-level (left) and observation-level (right) distributions together with the results of the resampling algorithms, for $N_{bins}(gen)=10$ and $N_{bins}(reco)=20$. In the truth-level space, the true distribution is shown together with the random sample that yields the best comparison with the observation-level distribution. The unfolded distribution using the iterative D'Agostini method is also shown. In the observation-level space, the smeared data distribution and the folded random sample distribution that yields the best comparison with the observation-level distribution.
  }
  \label{fig:fold_reco10gen5}
\end{figure}

\begin{figure}[!h]
  \centering
  \includegraphics[width=0.49\linewidth]{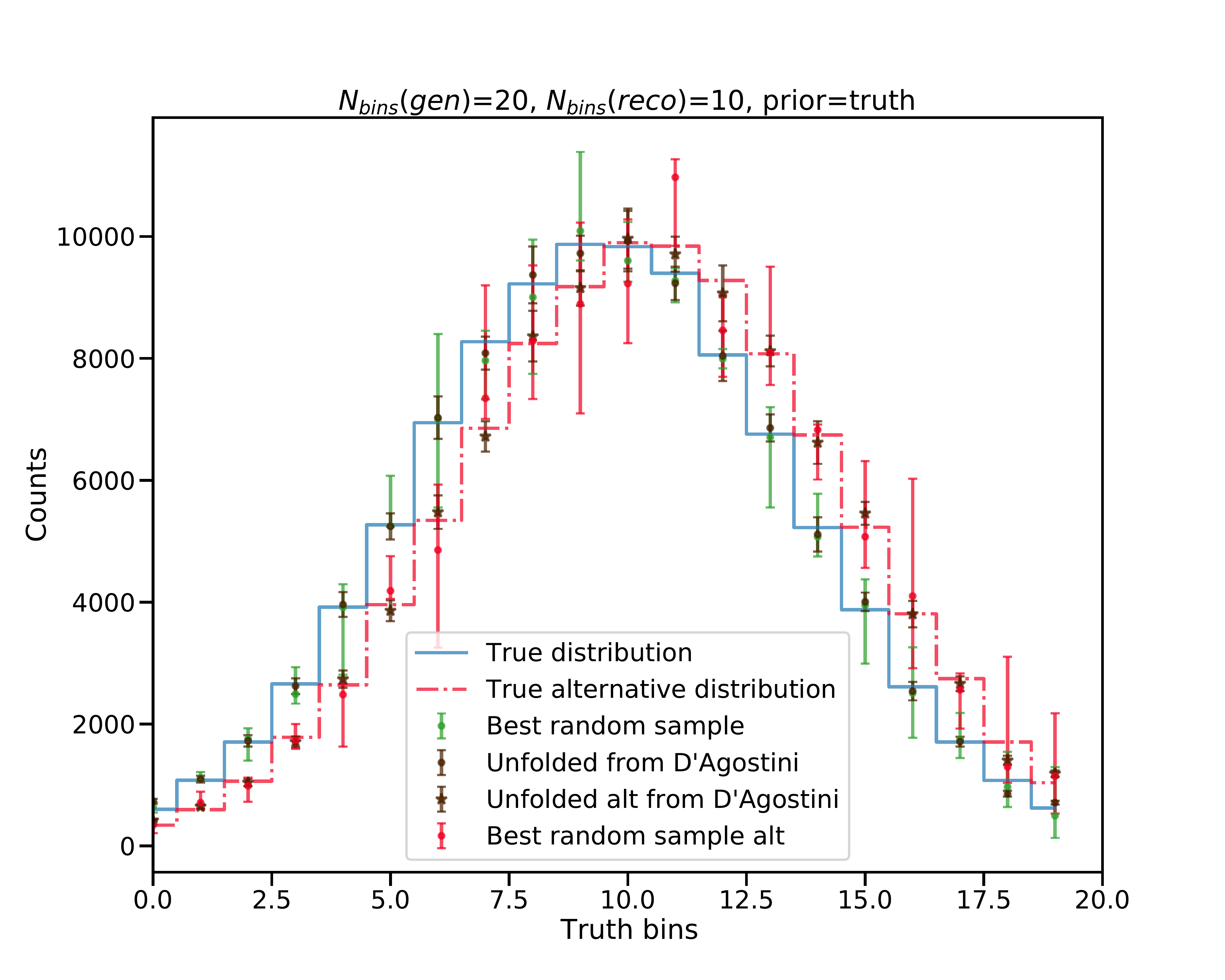}
  \includegraphics[width=0.49\linewidth]{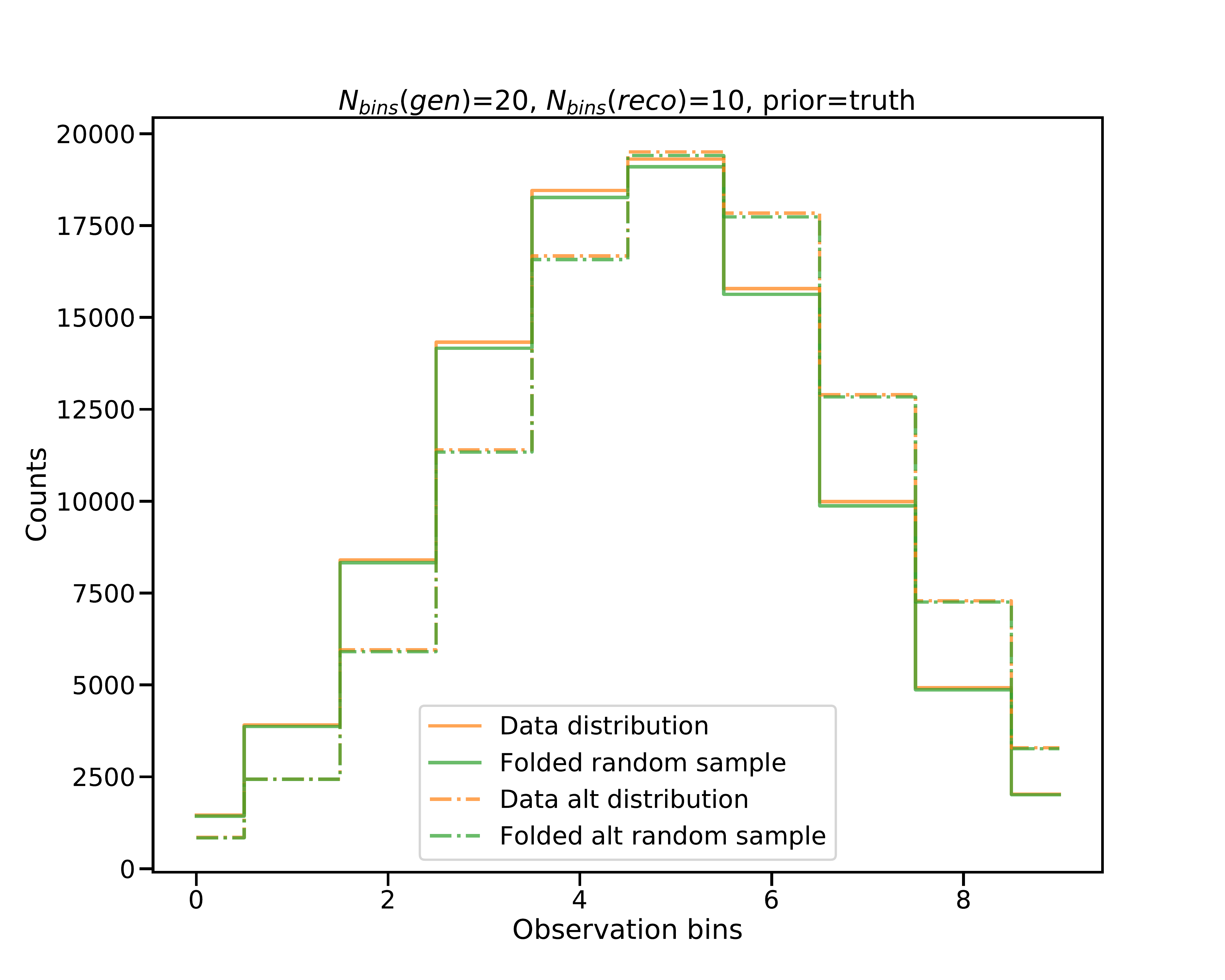}
  \caption{The truth-level (left) and observation-level (right) distributions together with the results of the resampling algorithms, for $N_{bins}(gen)=20$ and $N_{bins}(reco)=10$. In the truth-level space, the true distribution is shown together with the random sample that yields the best comparison with the observation-level distribution. The unfolded distribution using the iterative D'Agostini method is also shown. In the observation-level space, the smeared data distribution and the folded random sample distribution that yields the best comparison with the observation-level distribution.
  }
  \label{fig:fold_reco5gen10}
\end{figure}

Figure~\ref{fig:fold_reco10gen10} (top left),~\ref{fig:fold_reco10gen5} (top left), and~\ref{fig:fold_reco5gen10} (top left) respectively for the three discreteness scenarios,
show the observation-level distribution and the folded random sample that yields the best comparison with the observation-level distribution.
The top right panels shows the relative deviation between the two distributions.
The minimization procedure ensures that the chosen best random sample, once folded, is extremely close to the reference observation,
either the one used to build the response matrix or the alternative one used to test the sensitivity of the algorithm to physics not described by the response matrix.
The distributions tend to agree very well in most cases, as highlighted by the top right panels.

The best random sample is then compared in Figure~\ref{fig:fold_reco10gen10} (bottom left),~\ref{fig:fold_reco10gen5} (bottom left), and~\ref{fig:fold_reco5gen10} (bottom left)
respectively for the three discreteness scenarios, to the truth-level space to the truth-level distribution.
An unfolded distribution obtained by applying the iterative D'Agostini method as a reference to assess the performance of Algorithm 1.
The deviations, relative to the truth level distribution, of the best random sample of the unfolded D'Agostini distribution are then compared in the bottom right panel.
The total uncertainty in the best random sample is comparable with the corresponding uncertainty in the unfolded D'Agostini distribution,
but the bias of the central value is always on average smaller that the D'Agostini one.
For the case where the inverse problem is ill-defined problem, for the chosen scheme $N_{bins}(gen)=20$ and $N_{bins}(reco)=10$ some fluctuations are observed in the result of Algorithm~\ref{algo:thealgo}, although the uncertainty is pretty large.

\begin{figure}[!h]
  \centering
  \includegraphics[width=0.49\linewidth]{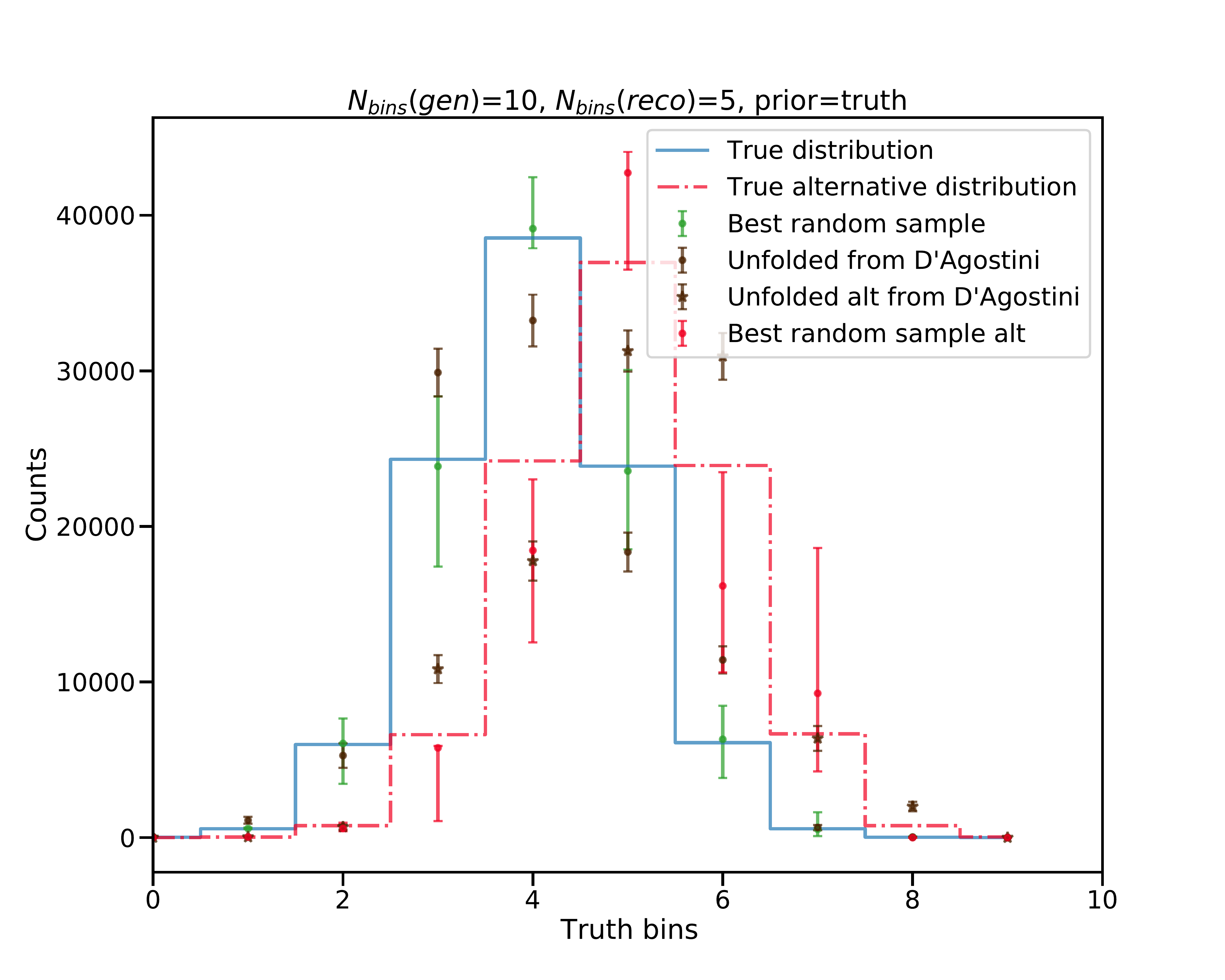}
  \includegraphics[width=0.49\linewidth]{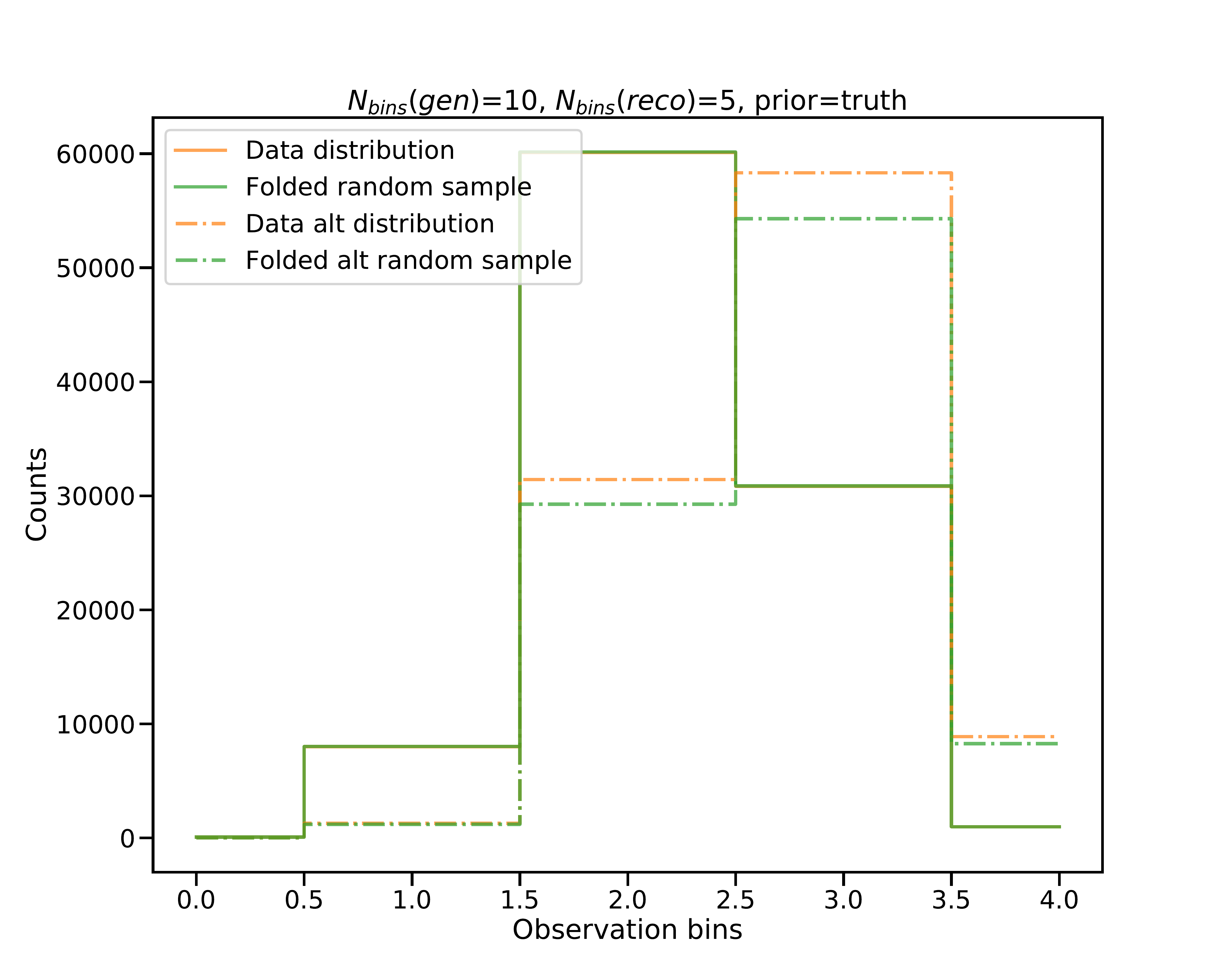}
  \caption{The truth-level (left) and observation-level (right) distributions together with the results of the resampling algorithms, for $N_{bins}(gen)=10$ and $N_{bins}(reco)=5$. In the truth-level space, the true distribution is shown together with the random sample that yields the best comparison with the observation-level distribution. The unfolded distribution using the iterative D'Agostini method is also shown. In the observation-level space, the smeared data distribution and the folded random sample distribution that yields the best comparison with the observation-level distribution.
  }
  \label{fig:fold_SMALLreco5gen10}
\end{figure}

In order to better investigate the behaviour in the pathological situation,
I generate a true distribution $\sim Gaus(4.5,1.0)$ and smear them with a random noise $\sim Gaus(1.0,0.4)$. The alternative spectrum is again obtained by a shift $\sim Gaus(1.0,0.3)$.
The results are shown in Figure~\ref{fig:fold_SMALLreco5gen10}, highlighting that
in this difficult situation---where the D'Agostini method somehow succumbs and is not able to suitably regularize the result---the Algorithm~\ref{algo:thealgo}
outperforms the standard unfolding both when ``unfolding'' the original spectrum and the alternative one.

In particular, the bias remains of the order of a few percent, whereas the bias for the D'Agostini method is more than one order of magnitude larger in some bins.
This indicates that Algorithm 1 may be particularly powerful when the other unfolding methods fail due to ill-definedness of the problem when $N_{bins}(gen)=20>N_{bins}(reco)$.

\begin{figure}[!h]
  \centering
  \includegraphics[width=0.49\linewidth]{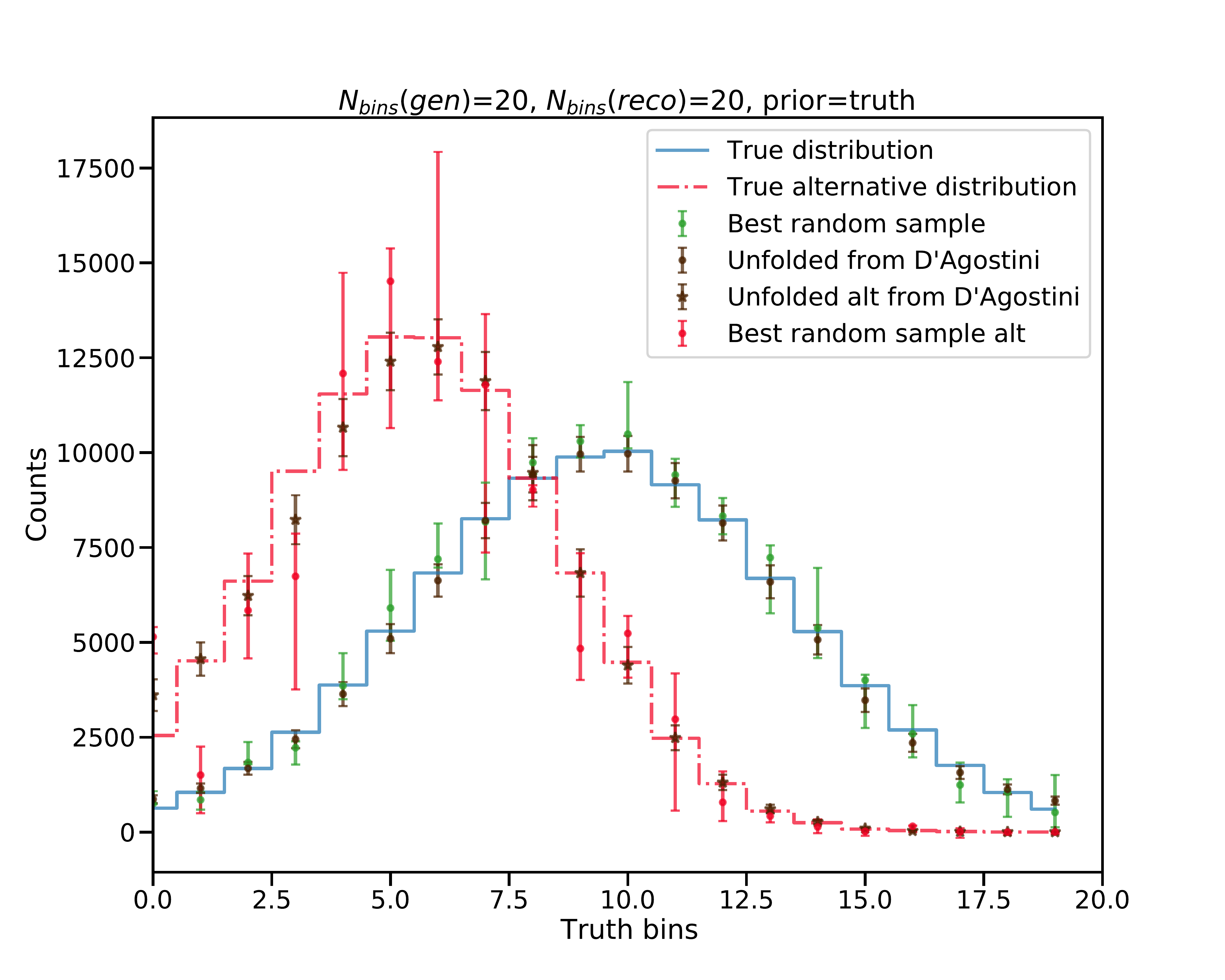}
  \includegraphics[width=0.49\linewidth]{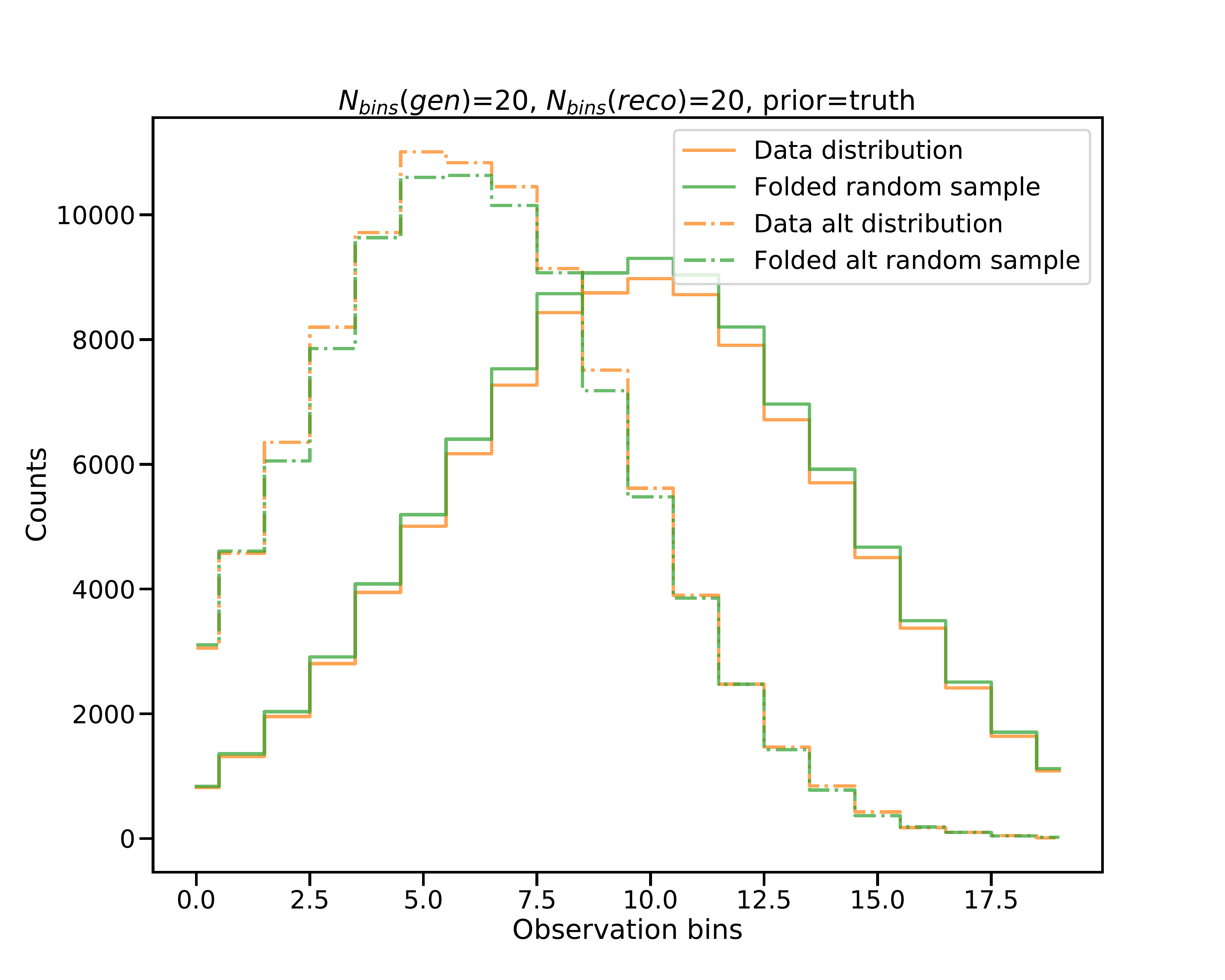}
  \caption{The truth-level (left) and observation-level (right) distributions together with the results of the resampling algorithms, for $N_{bins}(gen)=20$ and $N_{bins}(reco)=20$ and a remarkably non-diagonal smearing matrix. In the truth-level space, the true distribution is shown together with the random sample that yields the best comparison with the observation-level distribution. The unfolded distribution using the iterative D'Agostini method is also shown. In the observation-level space, the smeared data distribution and the folded random sample distribution that yields the best comparison with the observation-level distribution.
  }
  \label{fig:fold_NDEXTREMEreco10gen10}
\end{figure}

\begin{figure}[!h]
  \centering
  \includegraphics[width=0.49\linewidth]{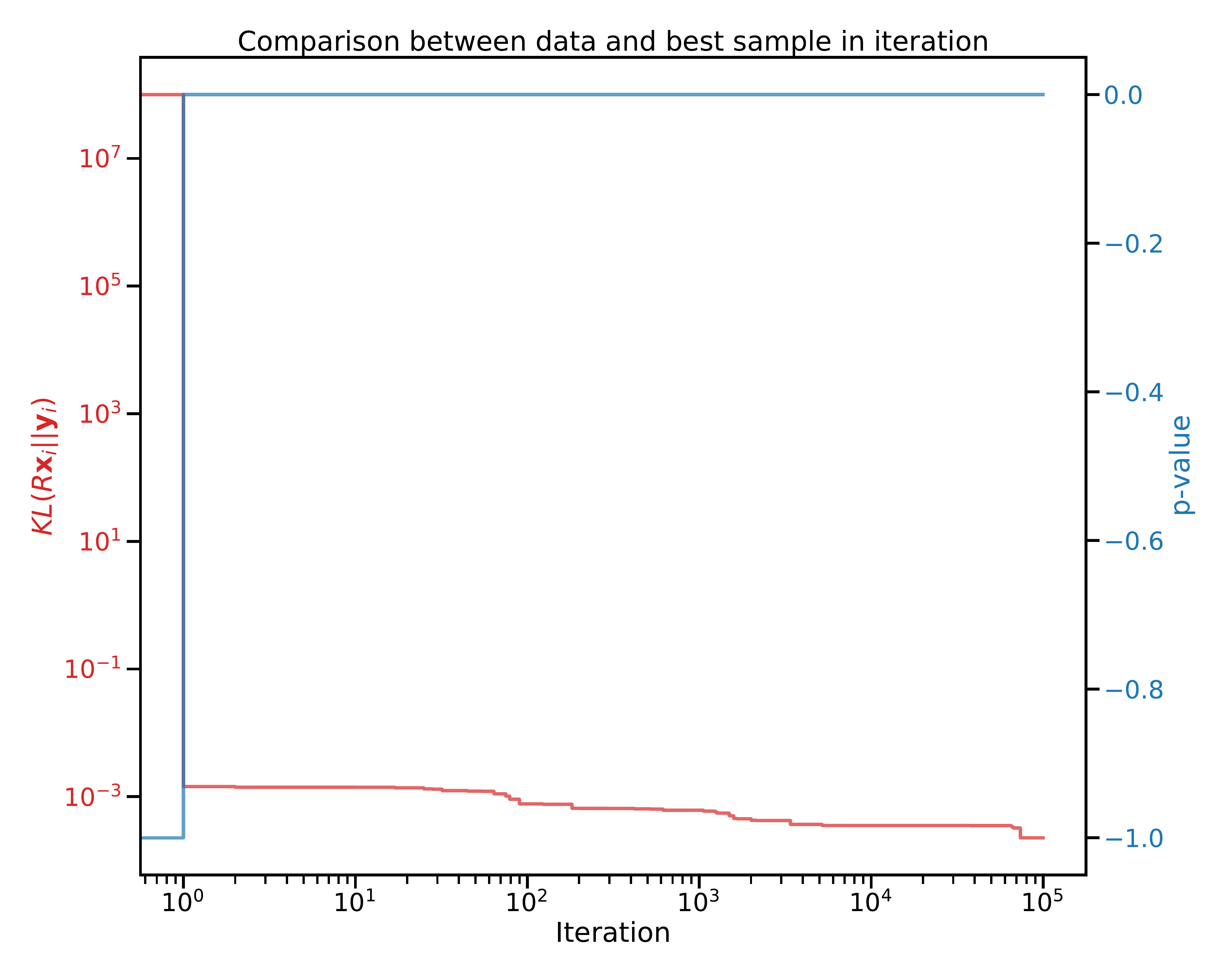}
  \caption{The $\chi^2$ test statistic (left vertical axis) and the $p$-value (right vertical axis) from a two-sample test between the folded best random sample and the observation-level distribution, plotted as a function of the number of random samples tested (\textit{iteration}), for $N_{bins}(gen)=20$ and $N_{bins}(reco)=20$.}
  \label{fig:iteration_reco10gen10}
\end{figure}

\begin{figure}[!h]
  \centering
  \includegraphics[width=0.49\linewidth]{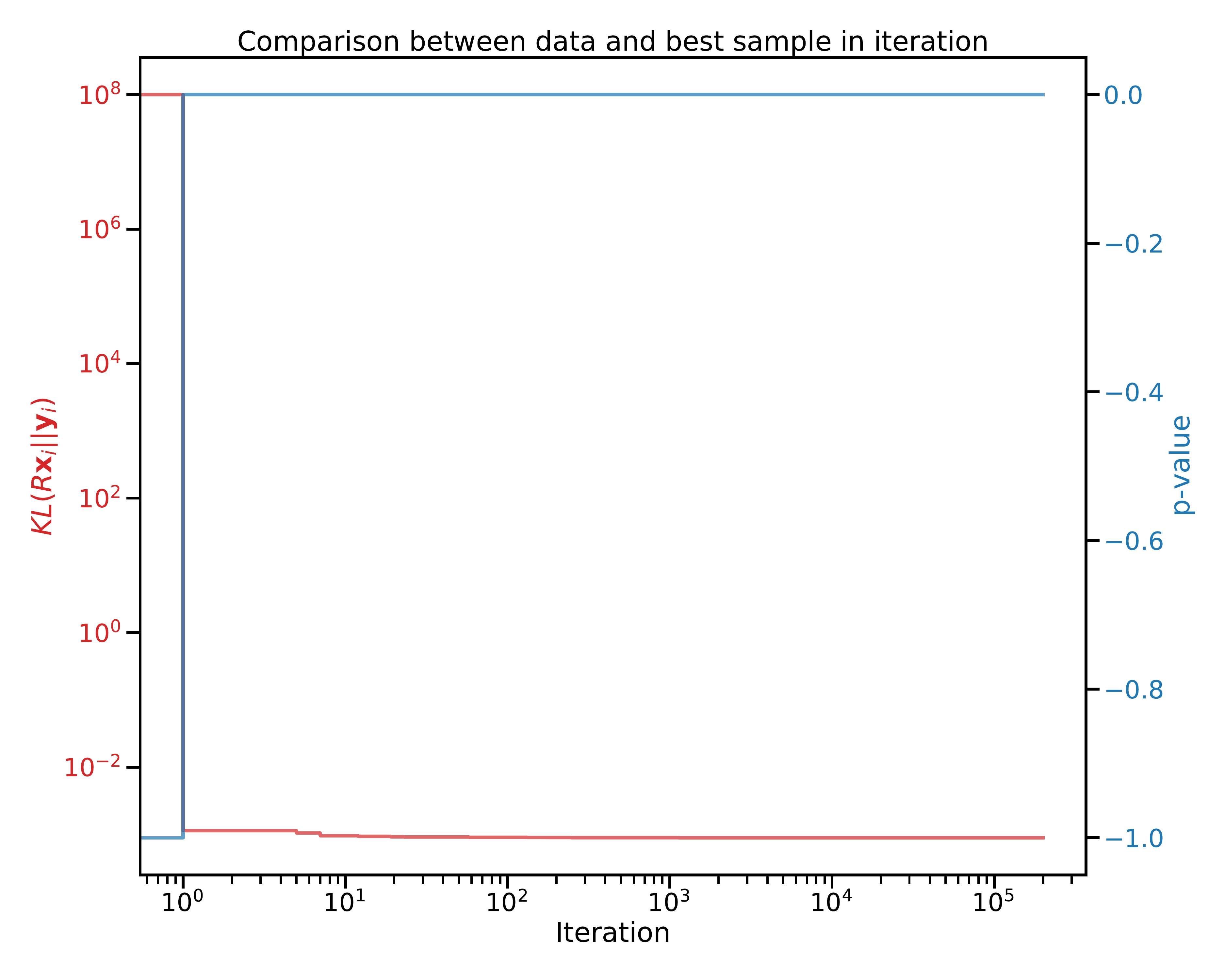}
  \caption{The $\chi^2$ test statistic (left vertical axis) and the $p$-value (right vertical axis) from a two-sample test between the folded best random sample and the observation-level distribution, plotted as a function of the number of random samples tested (\textit{iteration}), for $N_{bins}(gen)=10$ and $N_{bins}(reco)=20$.}
  \label{fig:iteration_reco10gen5}
\end{figure}

\begin{figure}[!h]
  \centering
  \includegraphics[width=0.49\linewidth]{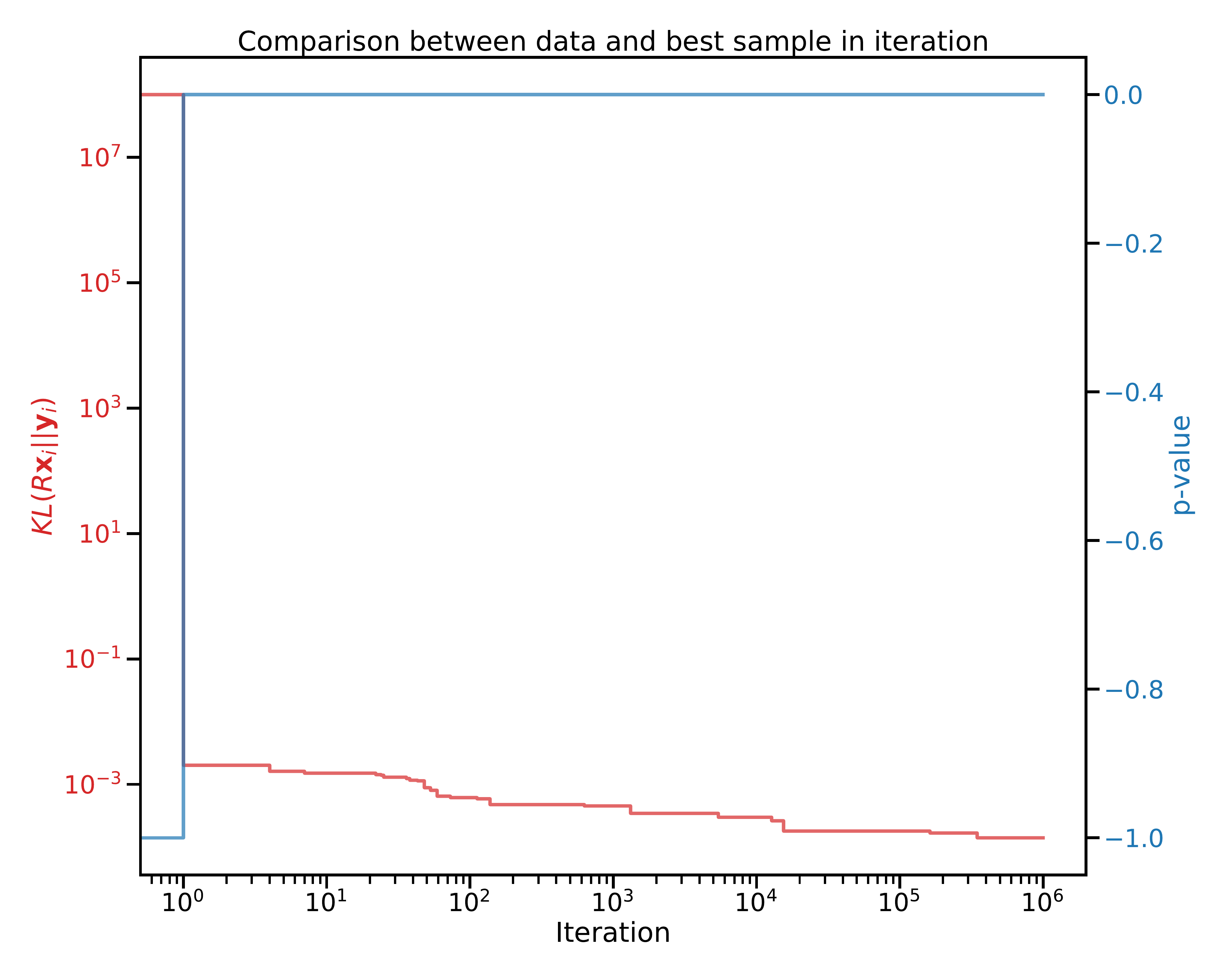}
  \caption{The $\chi^2$ test statistic (left vertical axis) and the $p$-value (right vertical axis) from a two-sample test between the folded best random sample and the observation-level distribution, plotted as a function of the number of random samples tested (\textit{iteration}), for $N_{bins}(gen)=20$ and $N_{bins}(reco)=10$.}
  \label{fig:iteration_reco5gen10}
\end{figure}

The convergence of the algorithm to the best random sample is illustrated in
Figures~\ref{fig:iteration_reco10gen10},~\ref{fig:iteration_reco10gen5},~\ref{fig:iteration_reco5gen10}, and~\ref{fig:fold_NDEXTREMEreco10gen10} for the three discreteness scenarios, respectively.

Coverage studies are yet to be performed.

\subsection{Dependence on the number of iterations}

The current configuration runs between $10^5$ and $2\times10^6$ iterations, but in the worst cases the convergence happens always within about $10^4$ iterations.
The production version of this algorithm would run without a specific number of iterations, and would just stop when the conditions on the two-sample $\chi^2$ test between the data and the folded best sample are satisfied.

\subsection{Dependence on the starting point for the sampling}

\begin{figure}[!h]
  \centering
  \includegraphics[width=0.49\linewidth]{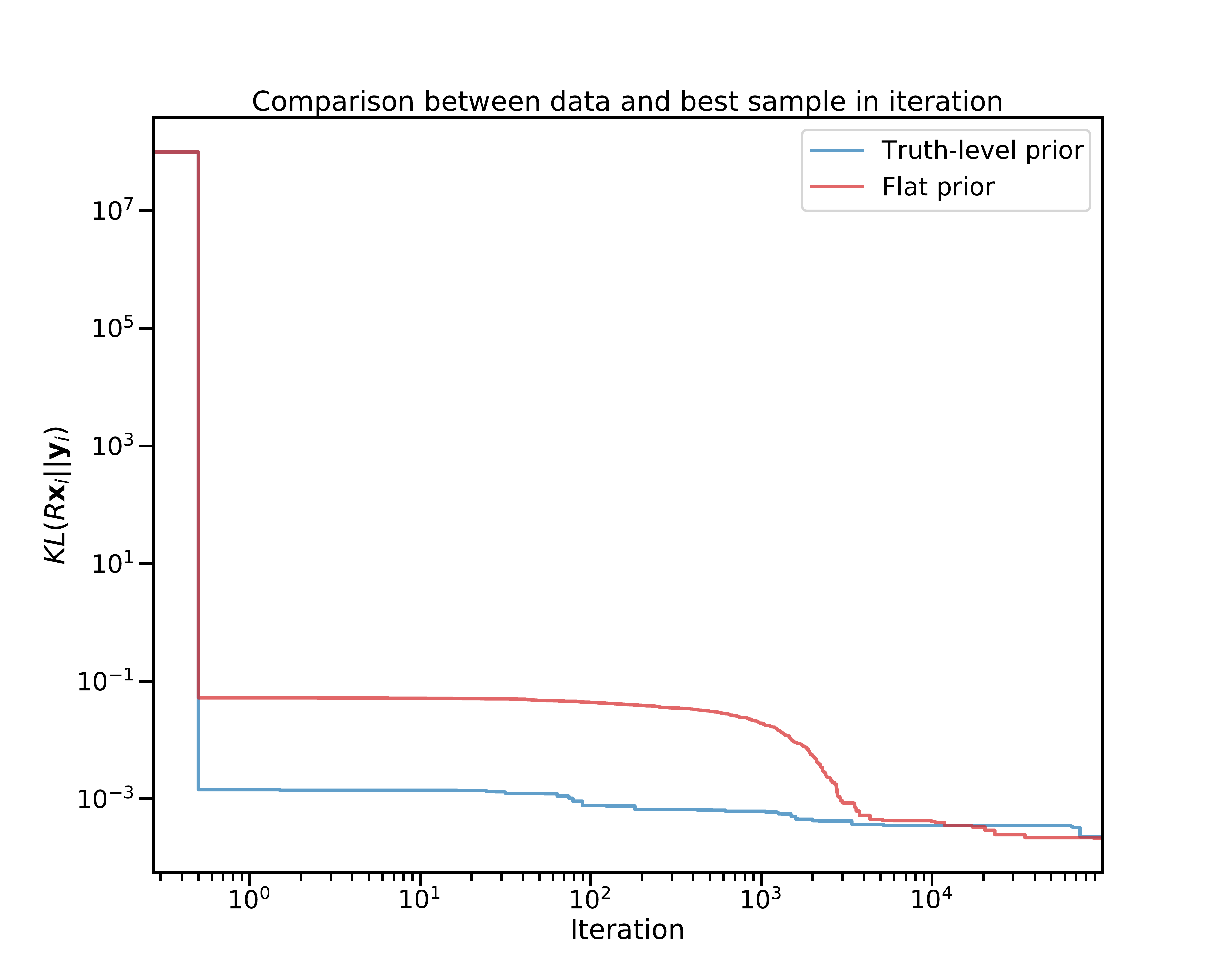}
  \includegraphics[width=0.49\linewidth]{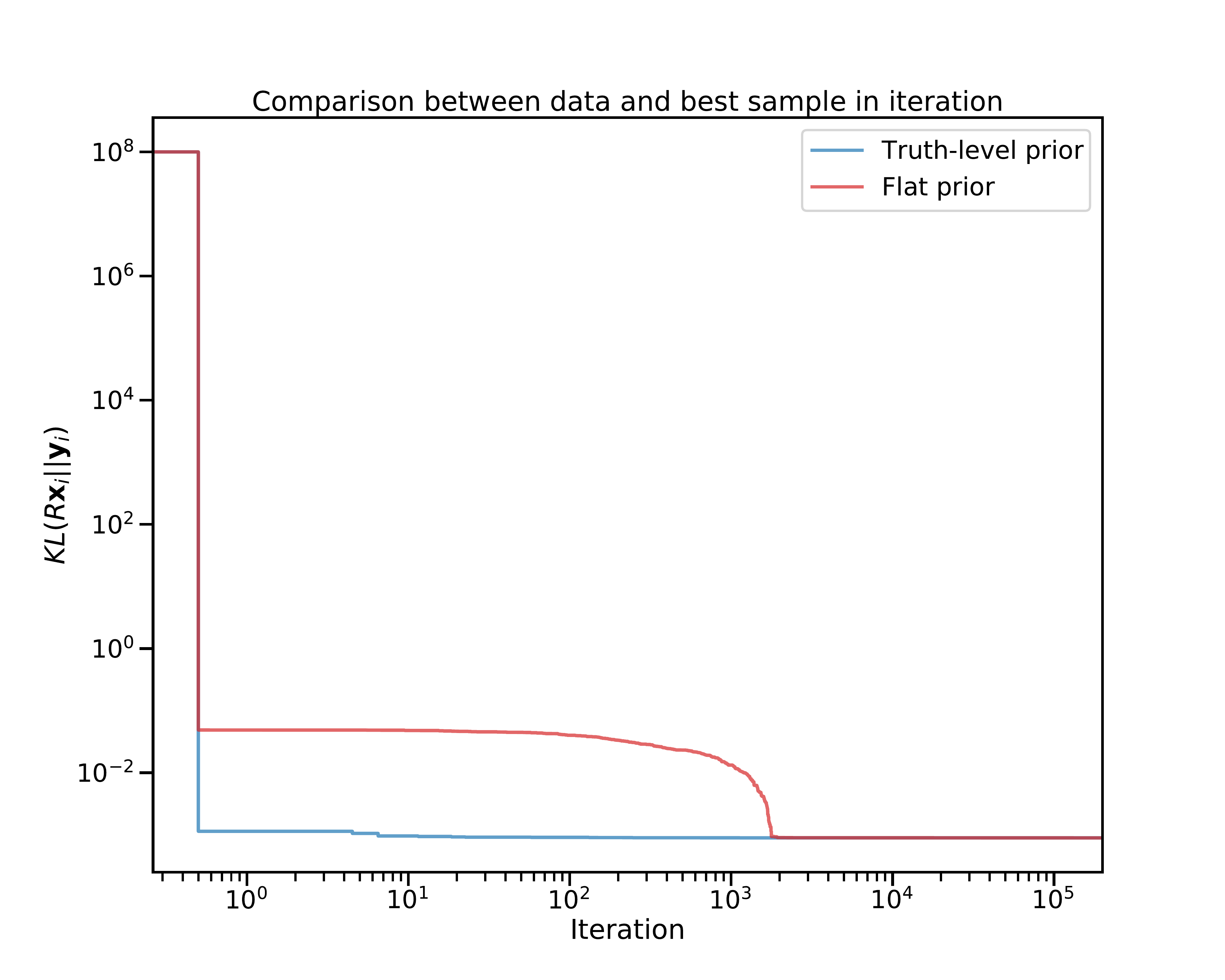}\\
  \includegraphics[width=0.49\linewidth]{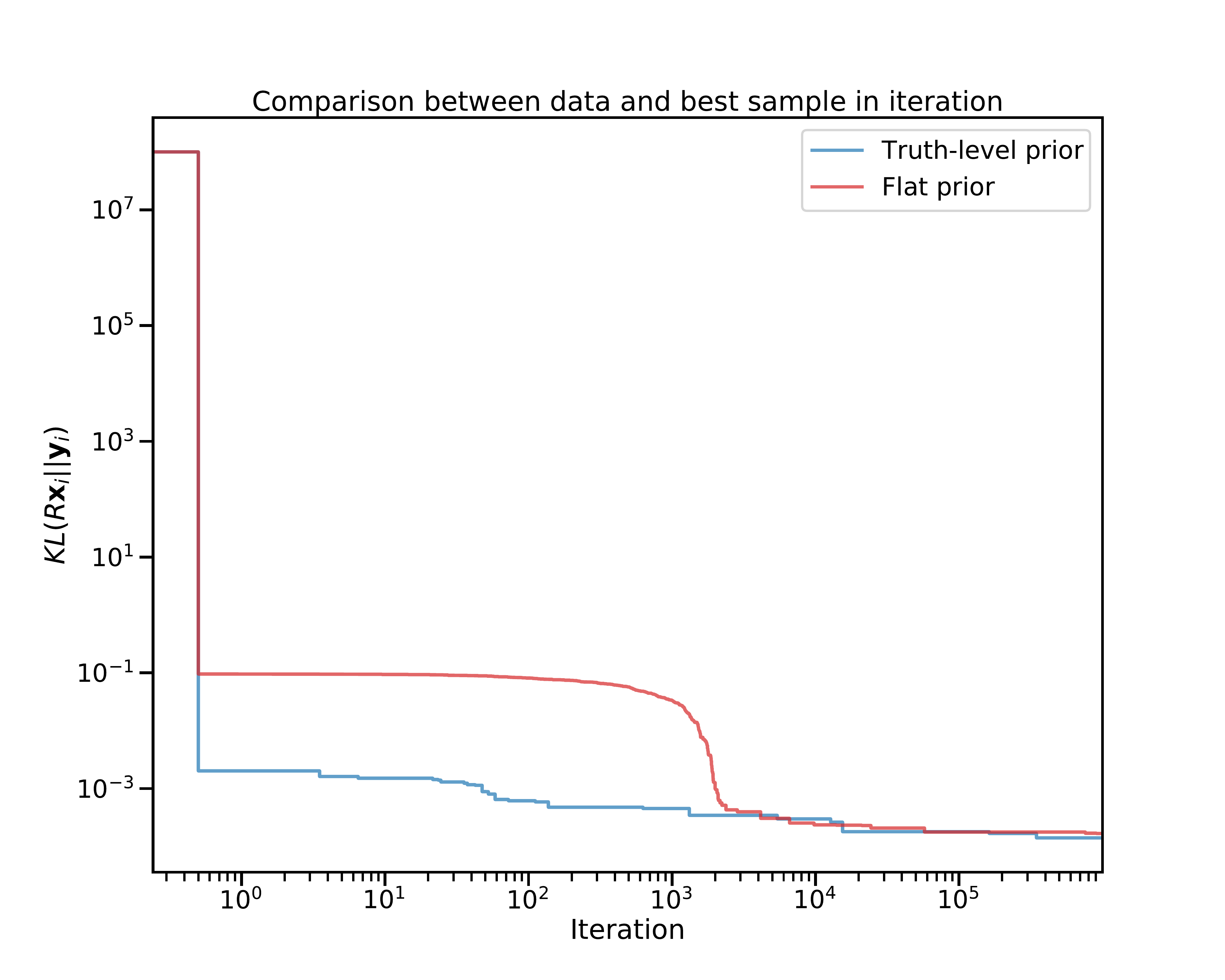}
 \caption{The evolution of the $\chi^2$ test statistic from a two-sample test between the folded best random sample and the observation-level distribution, as a function of the number of random samples tested (\textit{iteration}), for $N_{bins}(gen)=20$ and $N_{bins}(reco)=20$ (top left), for $N_{bins}(gen)=10$ and $N_{bins}(reco)=20$ (top right), and for $N_{bins}(gen)=20$ and $N_{bins}(reco)=10$ (bottom). Two choices of starting point for the sampling are shown.}
  \label{fig:flat_comparison}
\end{figure}

Currently the starting point for generating the random samples is the truth-level distribution.
On one side, when the data distribution agrees sufficiently with the folded version of the truth-distribution this can ensure a quicker convergence.
On the other side, when the data point to discrepancies with respect to the truth-level distribution, the convergence might not be so quick.
I therefore investigated different choices of starting point for this sampling algorithm.
The speed of the convergence to the good distribution is compared between the case when the prior is chosen as the truth-level distribution
and when it is chosen as a flat distribution with each bin set at half of the maximum value of the truth distribution:
this seems a good compromise between having a flat prior and allowing statistical uncertainties to have sufficient flexibility to be able to deform the shape
from a flat one to the one corresponding to the best agreement in the observation space.
Figure~\ref{fig:flat_comparison} shows the speed of convergence for the three $(N_{bins}(gen),N_{bins}(reco))$ scenarios $(20,20)$, $(10,20)$, and $(20,10$) respectively.
In all cases the convergence is slower at the beginning,
because sampling according to statistical fluctuations starting a flat shape takes time to start modelling a significantly non-flat distribution.
Once non-flatness is reached, however, the algorithm converges nicely to a best random sample whose folded version is within a few permille from the data distribution in each bin.
The 

Figures~\ref{fig:fold_reco10gen10} (top left),~\ref{fig:fold_reco10gen5} (top left), and~\ref{fig:fold_reco5gen10} (top left) respectively for the three discretensss scenarios,
show the observation-level distribution and the folded random sample that yields the best comparison with the observation-level distribution.
The top right panels shows the relative deviation between the two distributions.
The best random sample is then compared in Figure~\ref{fig:fold_reco10gen10} (bottom left) to the truth-level space to the truth-level distribution.
An unfolded distribution obtained by applying the iterative D'Agostini method as a reference to assess the performance of Algorithm 1.
The deviations, relative to the truth level distribution, of the best random sample of the unfolded D'Agostini distribution are then compared in the bottom right panel.

\begin{figure}[!h]
  \centering
  \includegraphics[width=0.49\linewidth]{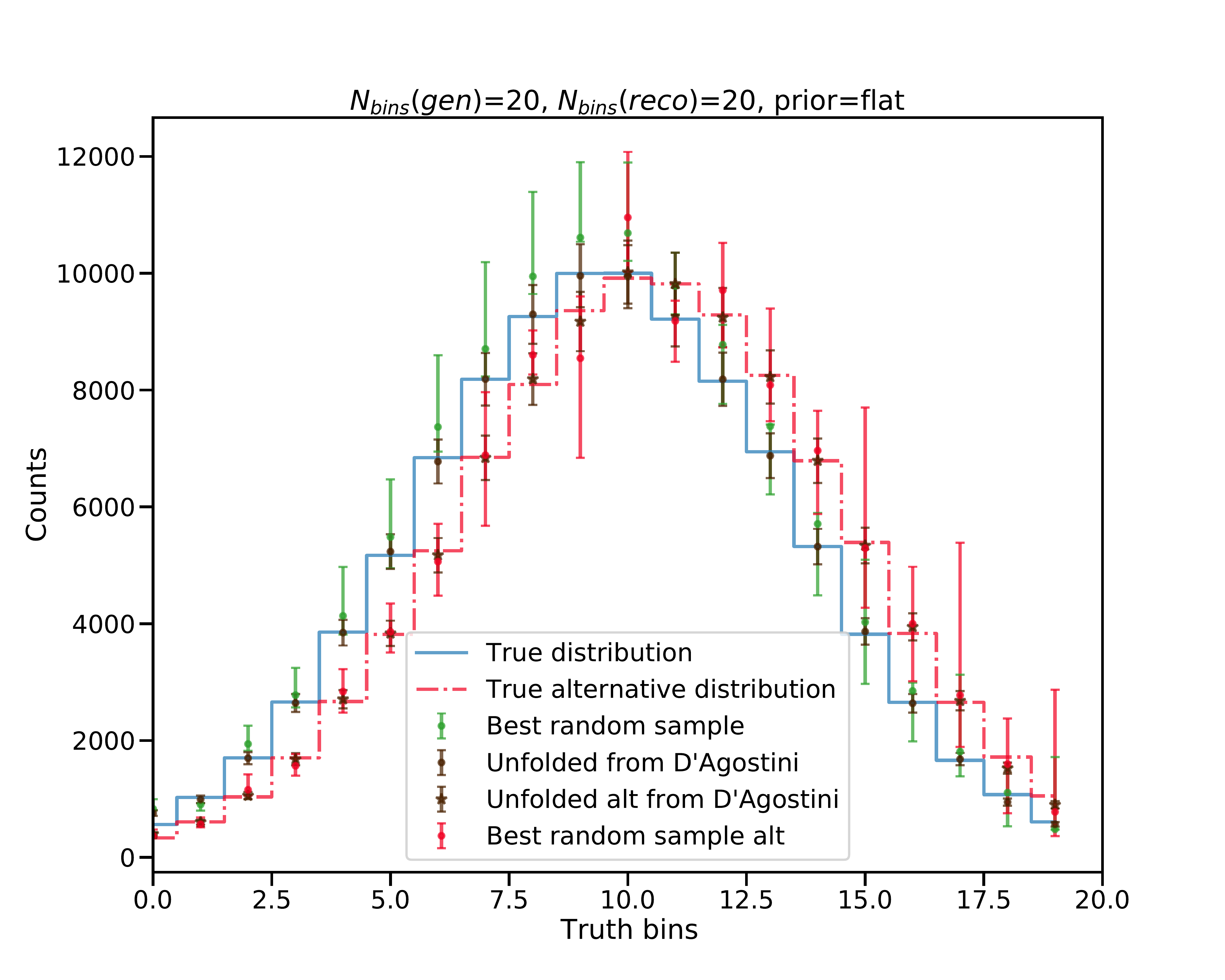}
  \includegraphics[width=0.49\linewidth]{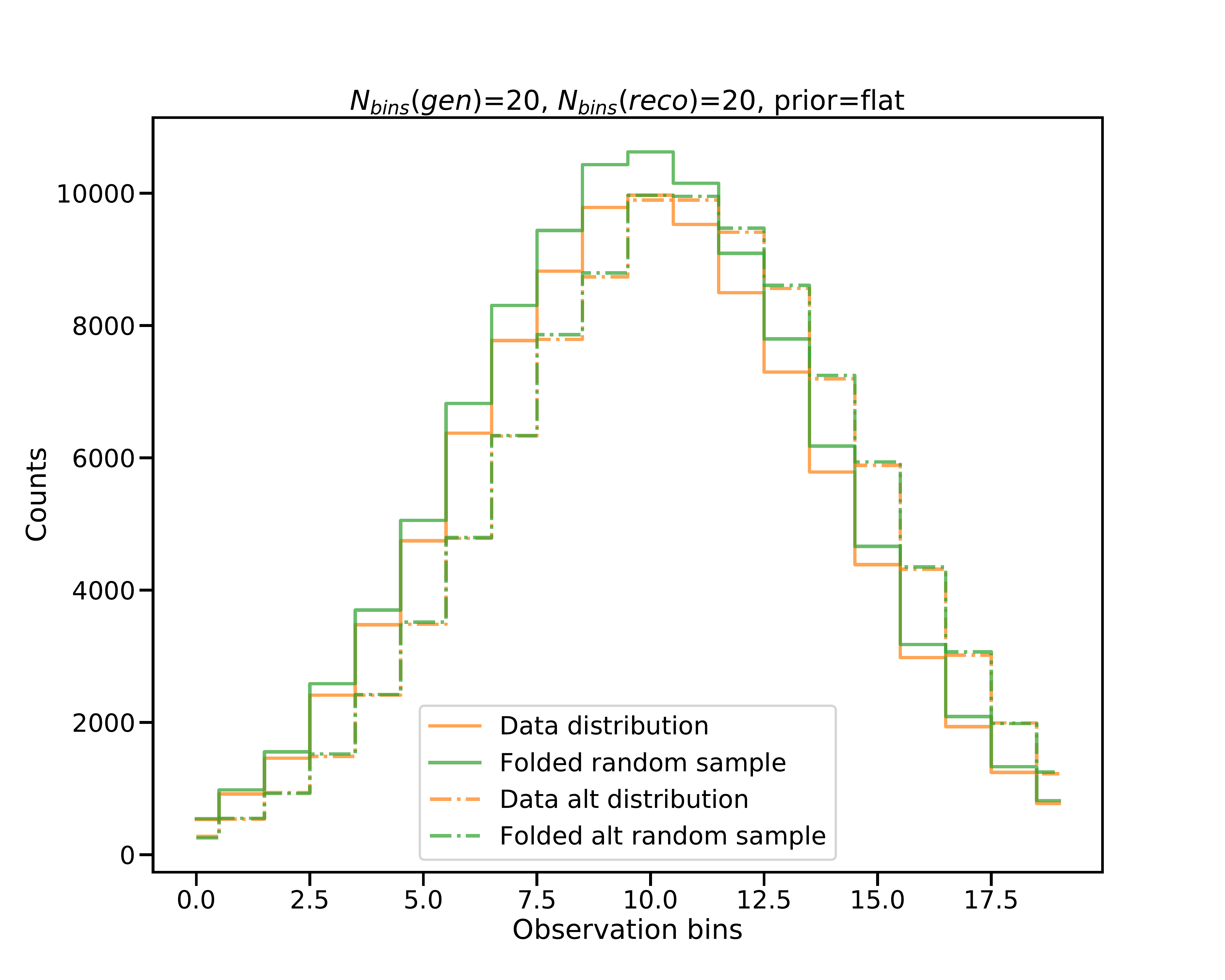}
  \caption{The truth-level (left) and observation-level (right) distributions together with the results of the resampling algorithms, for $N_{bins}(gen)=20$ and $N_{bins}(reco)=20$, for a flat starting point of the algorithm. In the truth-level space, the true distribution is shown together with the random sample that yields the best comparison with the observation-level distribution. The unfolded distribution using the iterative D'Agostini method is also shown. In the observation-level space, the smeared data distribution and the folded random sample distribution that yields the best comparison with the observation-level distribution.
  }
  \label{fig:fold_reco10gen10_flat}
\end{figure}

\begin{figure}[!h]
  \centering
  \includegraphics[width=0.49\linewidth]{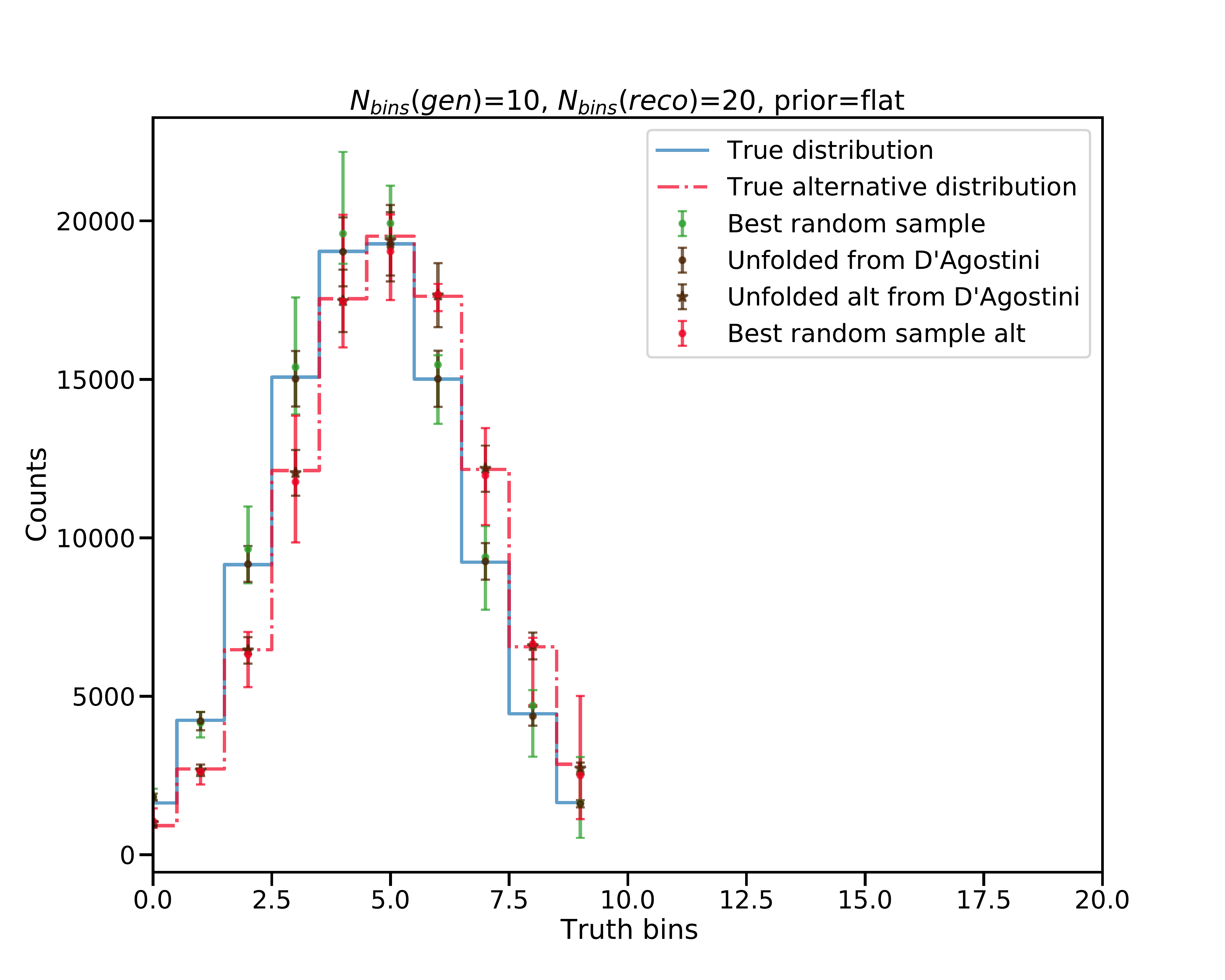}
  \includegraphics[width=0.49\linewidth]{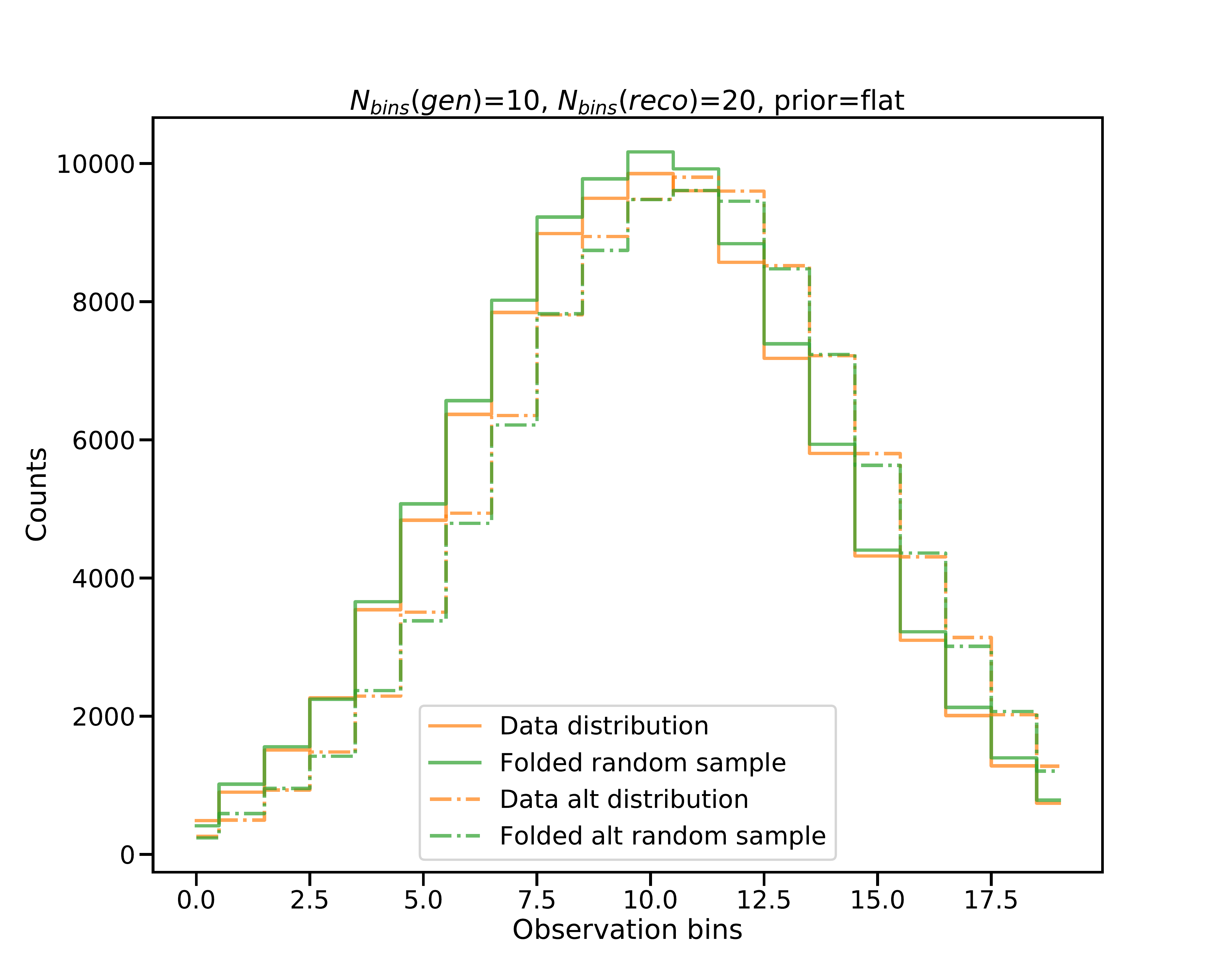}
  \caption{The truth-level (left) and observation-level (right) distributions together with the results of the resampling algorithms, for $N_{bins}(gen)=10$ and $N_{bins}(reco)=20$, for a flat starting point of the algorithm. In the truth-level space, the true distribution is shown together with the random sample that yields the best comparison with the observation-level distribution. The unfolded distribution using the iterative D'Agostini method is also shown. In the observation-level space, the smeared data distribution and the folded random sample distribution that yields the best comparison with the observation-level distribution.
  }
  \label{fig:fold_reco10gen5_flat}
\end{figure}

\begin{figure}[!h]
  \centering
  \includegraphics[width=0.49\linewidth]{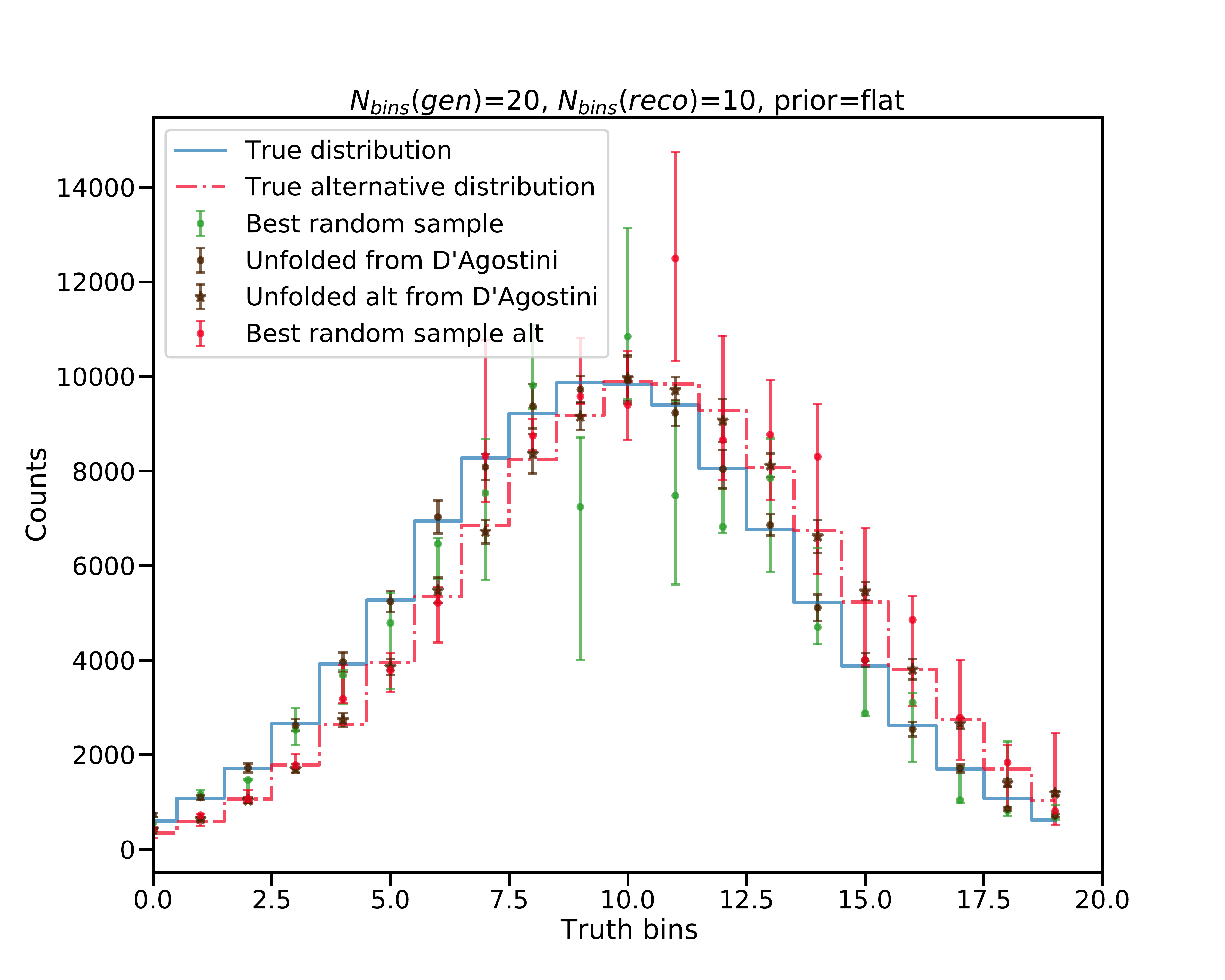}
  \includegraphics[width=0.49\linewidth]{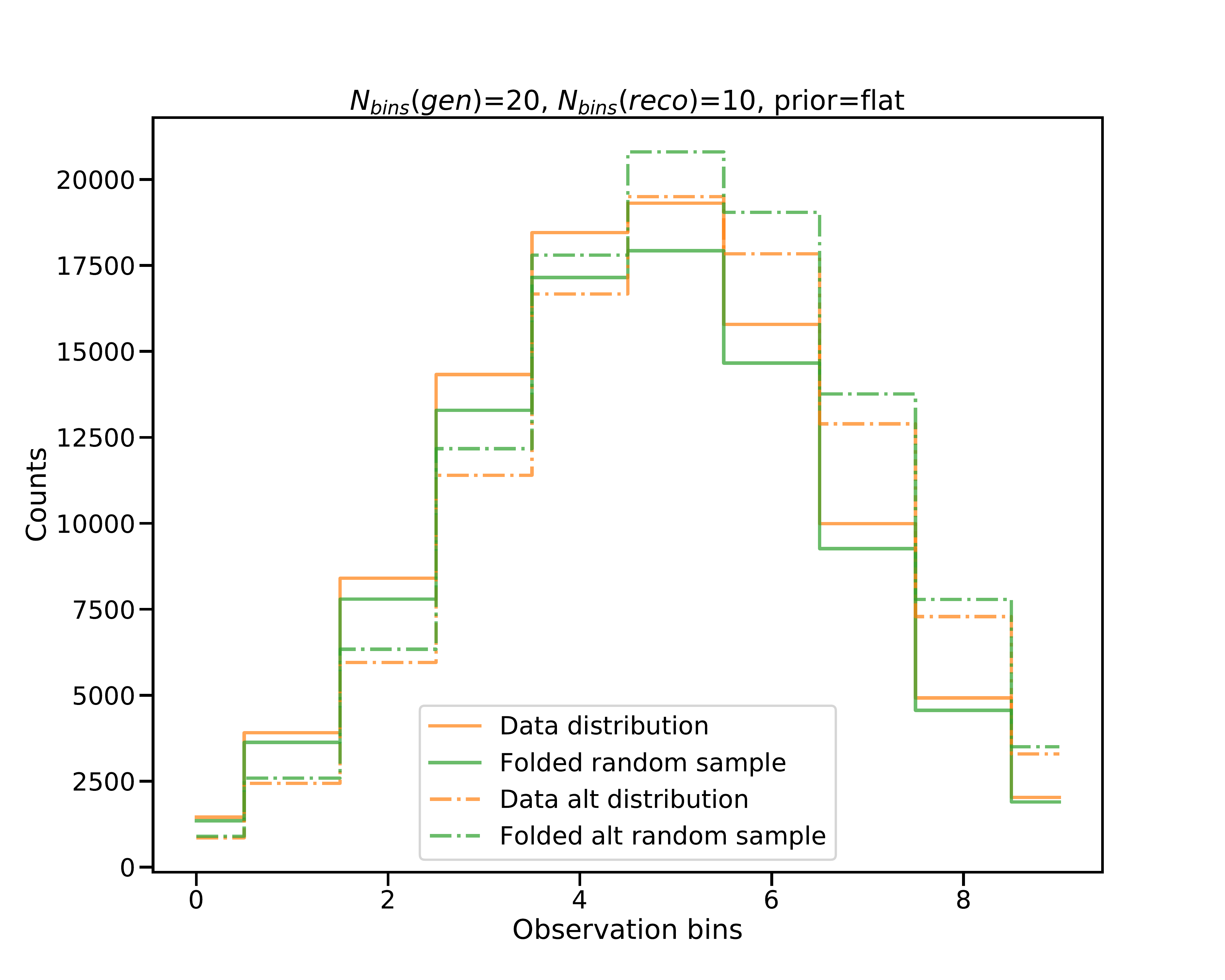}
  \caption{The truth-level (left) and observation-level (right) distributions together with the results of the resampling algorithms, for $N_{bins}(gen)=20$ and $N_{bins}(reco)=10$, for a flat starting point of the algorithm. In the truth-level space, the true distribution is shown together with the random sample that yields the best comparison with the observation-level distribution. The unfolded distribution using the iterative D'Agostini method is also shown. In the observation-level space, the smeared data distribution and the folded random sample distribution that yields the best comparison with the observation-level distribution.
  }
  \label{fig:fold_reco5gen10_flat}
\end{figure}

The best random sample is then compared in Figure~\ref{fig:fold_reco10gen10_flat} (down) to the truth-level space to the truth-level distribution, for $N_{bins}(gen)=20$ and $N_{bins}(reco)=20$.
An unfolded distribution obtained by applying the iterative D'Agostini method as a reference to assess the performance of Algorithm 1.
The deviations, relative to the truth level distribution, of the best random sample of the unfolded D'Agostini distribution are then compared in the bottom panel.

Figure~\ref{fig:fold_reco10gen5_flat} (top) shows, for $N_{bins}(gen)=10$ and $N_{bins}(reco)=20$,
the observation-level distribution and the folded random sample that yields the best comparison with the observation-level distribution.
The bottom panel shows the relative deviation between the two distributions.
For highly-populated distributions, the two-sample $\chi^2$ test is very sensitive to even small differences:
therefore, it is the best test ensuring that the chosen distribution, once folded, is extremely close to the observation.
Again the distributions agree to the permille level.

\subsection{Bottom-line test}

\begin{table*}[t]
  \topcaption{The bottom-line test: comparison between imperfect data and folded truth in folded space, and between best random from imperfect data and truth-level truth in unfolded space, for Algorithm 1 and for D'Agostini Unfolding. The $\chi^2/NDOF$ test statistic is reported. The regular NDOF is used, its computation neglecting the covariance matrix.}
  \label{tab:bottomline}
  \centering
  \scriptsize
  \begin{tabular}{lcccccc}
    \hline
    \multirow{3}{*}{Configuration} & \multicolumn{2}{c}{Algo1 $\chi^2/NDOF_{foldedSpace}$} & \multicolumn{2}{c}{Algo1 $\chi^2/NDOF_{unfoldedSpace}$} & \multicolumn{2}{c}{D'Agostini $\chi^2/NDOF_{unfoldedSpace}$}\\
    & \multicolumn{2}{c}{$(altData, foldedTruth)$} & \multicolumn{2}{c}{$(bestRandomFromAltData, truth)$} & \multicolumn{2}{c}{$(d'Agostini unfolded, truth)$} \\
    & init:truth & init:flat & init:truth & init:flat & init:truth & init:flat\\
    \hline
    $N_{bins}(gen)=20$, $N_{bins}(reco)=20$ & 280.753 & 280.753 & 306.196 & 322.679 & 307.210 & 307.210 \\
    $N_{bins}(gen)=10$, $N_{bins}(reco)=20$ & 285.260 & 285.260 & 592.488 & 596.413 & 603.189 & 603.189 \\
    $N_{bins}(gen)=20$, $N_{bins}(reco)=10$ & 542.366 & 542.366 & 302.102 & 358.552 & 304.772 & 304.772 \\
    \hline
  \end{tabular}
\end{table*}

The bottom-line test, suggested by Cousins and colleagues~\cite{Cousins:2016ksu}, can be expressed as \textit{the inference in the unfolded space should not be better than that in the folded space}.

In this contest, the test consists ensuring that the comparison in the folded space between the original and alternative data (in our HEP analogy,
the reconstructed simulated NLO distribution and the observed data that assume a NNLO truth) induces
the same inference as the comparison in the unfolded space between the original truth (NLO simulated truth) and the random distribution obtained ``unfolding''
the alternative data (the unfolded data pointing to a NNLO spectrum).
The test is performed for all the combinations described above (using either the generated spectrum or a flat spectrum as a starting point) and for the D'Agostini unfolding,
confirming that the procedure seems to generally pass the botton-line test, although probably more studies are needed accounting for the degrees of freedom lost in the unfolding procedure.
The results are anyway summarized in Table~\ref{tab:bottomline}

\section{Discussion}

Algorithm~\ref{algo:thealgo} uses the data distribution as a target for folding each resampled realization of the truth.
It assumes Poisson uncertainties in each bin to resample.
The truth distribution is used as a starting point, but tests with a flat distribution show that the performance is similar;
the algorithm seems therefore robust with respect to changes of the starting point of the resampling,
and seems not to be subject to any bias due to the truth distribution.
Starting from a distribution far from the expected result seems to require just more iterations to convergence in the folded space.
Even tests done by using as a target data that come from a shifted version of the true spectrum show that the ``unfolded'' result
matches with the corresponding underlying true spectrum rather than with the spectrum used as a starting point or that used
to build the response matrix.
This is important, in that it means that in all applications one can safely use existing Monte Carlo simulated spectra as a starting
point without worrying about introducing biases to the assumed starting point distribution.
This is a very desirable property of this algorithm.

Bias caused by the non-diagonality of the response matrix is instead unavoidable, and warrants regularization methods:
preliminar studies indicate that imposing smoothness conditions on the sampled distribution in true space results in too strong a regularization,
whereas the procedure recommended by Kuusela~\cite{kuusela} of unfolding with finer binning and then rebin the unfolded result seems applicable with profit to this method.

Of course the algorithm is still dependent on the assumed truth distribution via the response matrix,
and this probably explains the imperfections in the response when applied to different data.
This is however probably unavoidable (after all, we need a model for the detector response, although smarter things maybe could be done).

Uncertainties in the response matrix induce non-negligible uncertainties in the best candidate truth distribution, but the estimator seems unbiased in most cases.

When traditional unfolding fails (e.g. when $N_{bins}(reco)<N_{bins}(gen)$, this resampling-based algorithm retains its full performance,
and seems very competitive when traditional unfolding algorithm have a poor performance.

Care should be taken in the low-yields regime to employ suitable alternative to the Pearson $\chi^2$ test statistic.
I have described a few alternative choices and their performance in Section~\ref{sec:statTest}: the public version of the
code will include the appropriate software switches to enable all these alternative choices.

Further developments include an ongoing refined treatment of statistical and systematic uncertainties via resampling from a joint likelihood
built from the simulated spectrum used to build the response matrix, in a pseudo-ABC way as exemplified by Algorithm~\ref{algo:algoabc}.

\section{Summary}

Matrix inversion problems have many application in experimental particle physics, where it is often referred to as the \textit{unfolding problem}.
The response of a detector is modelled via a matrix that connects the observed spectra and the underlying true spectra.
Traditional unfolding algorithms resolve the inverse problem by inverting the matrix;
they tend to deal with the cases where the inversion is tricky by introducing regularization term in the inversion procedure.
The regularization requires some assumptions on the true unfolded solution, that generally result in a bias towards the assumed spectrum.

In this Manuscript, I have proposed an algorithm that avoids an explicit matrix inversion by resampling in the true space.
A large number of random spectra are drawin in truth space and folded until the folded realizations are in excellent agreement with the target data,
using stochastic optimization to reach the desired target.
The sampling is performed on the discrete problem (sampling from the expected bin contents), resulting in an implicit degree of regularization.
Using the data as a target avoids other biases towards true spectrum: the starting point for the iterative resampling seems indeed to not influence the result,
neither when starting from a dramatically different spectrum (flat) nor when starting from a close-but-different spectrum (the typical situation of NLO-vs-NNLO spectra in particle physics).
Biases induced by the non-diagonality of the problem remain significant when the response matrix is significantly non-diagonal, and further regularization must be applied.
Avoiding the inversion of the response matrix has also the important perk of avoid issue that arise when an ill-define response matrix ($N_{bins}(gen)>N_{bins}(reco)$ is inverted.
The algorithm is shown to be able to outperform standard unfolding algorithms in these pathological situations.

Further improvements include a procedure for resampling from a likelihood that models the response matrix and its uncertainties, in a way which resembles some variants of Approximate Bayesian Computation.

I will update this Manuscript with additional bias and coverage studies.

\section*{Acknowledgements}
I wish to thank Robert Cousins, Tommaso Dorigo, Mikael Kuusela, and Igor Volobouev for their feedback on earlier versions of the idea of unfolding-by-folding,
the CMS Statistics Committee for the several profound discussions on the problem of unfolding spectra,
and the CMS Collaboration for providing a stimulating landscape for unfolding problems.

I thank the Center for Cosmology, Particle Physics, and Cosmology, within the Institut de recherche en Math\'ematique et Physique of the Universit\'e catholique de Louvain,
for the independence it gives me in the pursuit of my research interests. I also thank the CP3 and CECI computing clusters for providing the computing resources that have been crucial to this project.

This work was supported by the Universit\'e catholique de Louvain, Belgium, under the FSR grant \texttt{IRMPFSR19MOVELem}.

\bibliographystyle{splncs03_unsrt}
\bibliography{bibliography}

\end{document}